\documentclass[pdflatex,sn-mathphys-num]{sn-jnl}% Math and Physical Sciences Numbered Reference Style
\usepackage{graphicx}%
\usepackage{multirow}%
\usepackage{amsmath,amssymb,amsfonts}%
\usepackage{amsthm}%
\usepackage{mathrsfs}%
\usepackage[title]{appendix}%
\usepackage{xcolor}%
\usepackage{textcomp}%
\usepackage{manyfoot}%
\usepackage{booktabs}%
\usepackage{algorithm}%
\usepackage{algorithmicx}%
\usepackage{algpseudocode}%
\usepackage{listings}%
\theoremstyle{thmstyleone}%
\theoremstyle{thmstyletwo}%

\theoremstyle{thmstylethree}%

\begin{document}

\title[Article Title]{A Hybrid Multi-Goal Path Planning Framework for Powered Wheelchairs in Dynamic Environments: From Simulation to Real-World Validation}

%%=============================================================%%
%% GivenName	-> \fnm{Joergen W.}
%% Particle	-> \spfx{van der} -> surname prefix
%% FamilyName	-> \sur{Ploeg}
%% Suffix	-> \sfx{IV}
%% \author*[1,2]{\fnm{Joergen W.} \spfx{van der} \sur{Ploeg} 
%%  \sfx{IV}}\email{iauthor@gmail.com}
%%=============================================================%%

\author*[1,2]{\fnm{Damla} \sur{Leblebicio\u{g}lu}}\email{leblebicioglu.d@northeastern.edu}

\author[1,2]{\fnm{Ege} \sur{\"{O}zg\"{u}l}}\email{ozgul.e@northeastern.edu}

\author[1,2]{\fnm{Ta\c{s}k{\i}n} \sur{Pad{\i}r}}\email{t.padir@northeastern.edu}  

\affil[1]{\orgdiv{Electrical and Computer Engineering}, \orgname{Northeastern University}, \orgaddress{\street{360 Huntington Ave}, \city{Boston}, \postcode{02115}, \state{MA}, \country{USA}}}

\affil[2]{\orgdiv{Institute for Experiential Robotics}, \orgname{Northeastern University}, \orgaddress{\street{360 Huntington Ave}, \city{Boston}, \postcode{02115}, \state{MA}, \country{USA}}}

%%==================================%%
%% Sample for unstructured abstract %%
%%==================================%%

\abstract{Autonomous powered wheelchairs (PWs) operating in human-centered environments must combine efficient multi-destination path planning with continuous collision avoidance strategies in the presence of static and dynamic obstacles. This study presents a hybrid framework integrating Traveling Salesman Problem (TSP)-based global task ordering, Potential-Field (PF) guidance, and local Rapidly-exploring Random Tree (RRT) and A-star (A*) planners. PF guidance transfers global path information to local planners, helping the robot stay aligned with the global path during obstacle avoidance. Local planning is enhanced with direction-aware, extended-neighborhood A* or RRT using environment-aware sampling and two-step extension. The TSP--PF--RRT and TSP--PF--A* methods are evaluated in normal, crowded, and extra-crowded simulation environments using point-based and elliptical robot models. The elliptical model accounts for platform dimensions during collision avoidance phase. Hardware tests on the physical PW use motion capture (MoCap) for ground-truth measurements, while outside-MoCap experiments rely on onboard LiDAR and RGB-D sensing. Results show that incorporating the wheelchair geometry changes the observed collision behavior, yielding comparable collision avoidance performance between the two hybrid planners under the elliptical model.}


%%\pacs[JEL Classification]{D8, H51}

%%\pacs[MSC Classification]{35A01, 65L10, 65L12, 65L20, 65L70}

\maketitle

\section{Introduction}\label{sec1}

\begin{figure}[t]
\centering
\includegraphics[width=1.0\textwidth]{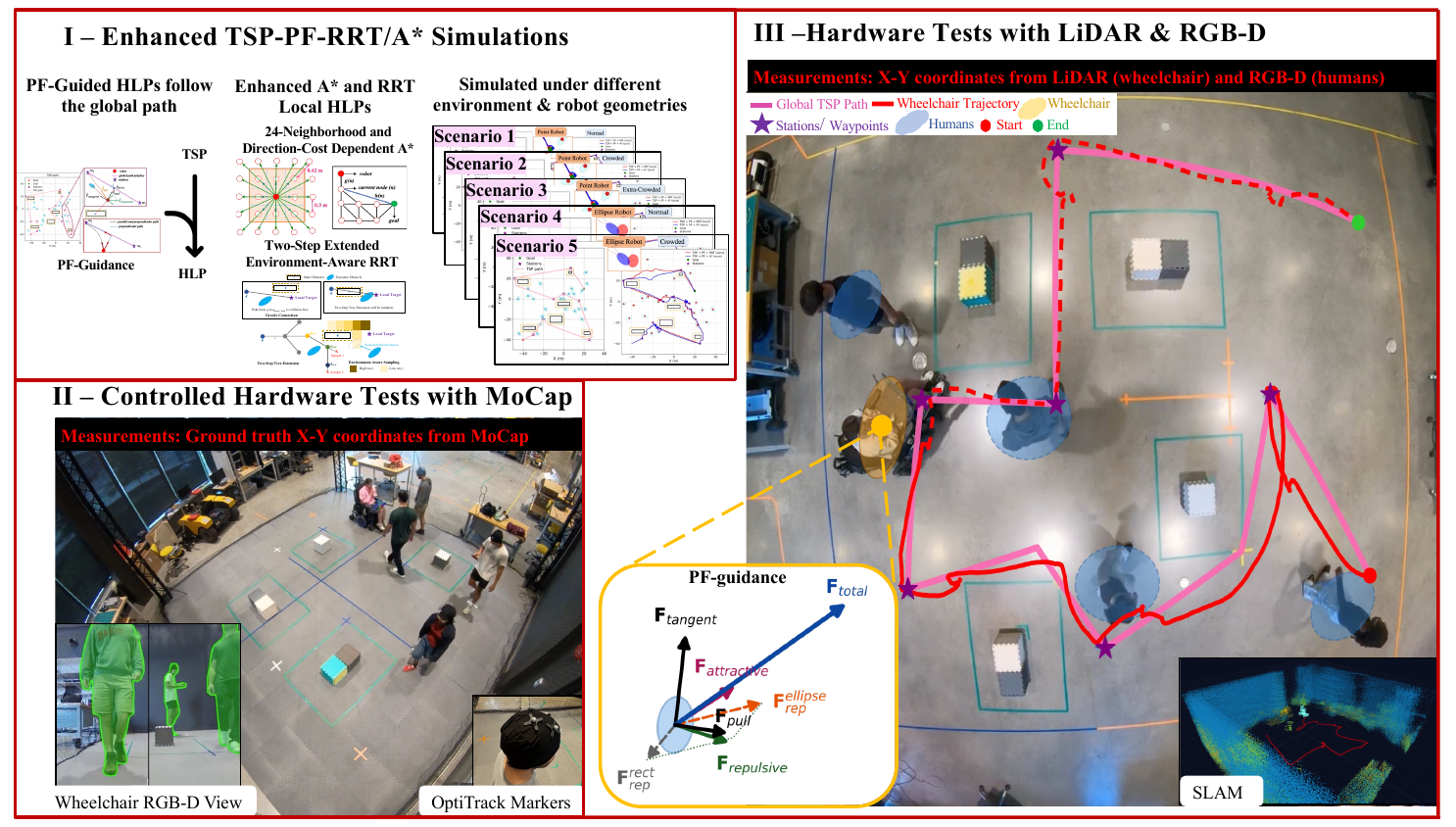}
\caption{Overview of the proposed framework. (I) Simulation-based evaluation of the enhanced TSP--PF--RRT and TSP--PF--A* planners with point and elliptical wheelchair models in environments of increasing density. (II) Controlled hardware experiments with MoCap ground truth. (III) Hardware experiments with onboard LiDAR and RGB-D sensing; the inset shows the PF forces acting on the wheelchair (magnitudes are schematic).}
\label{fig:teaser}
\end{figure}

%Paragraph1: formulation of the problem
PWs can play an important role in improving independent mobility and quality of life for older adults and individuals with severe motor impairments, including those associated with neurological conditions such as amyotrophic lateral sclerosis (ALS) and multiple sclerosis (MS)~\cite{simpson2008many, padir2015towards}. However, enabling a robotic wheelchair to operate autonomously in human-centered environments remains challenging due to dynamic obstacles, narrow passages, environmental perception uncertainties, and crowded spaces in which safe collision-free motion must be maintained~\cite{loke2026minimal,simpson2008many, sinyukov2019wheelchairs}. These challenges motivate continued research toward safer, more intelligent, and practically deployable powered wheelchair systems for everyday use~\cite{cooper2008trends}. In many practical applications, the path-planning problem is also more complex than traveling from a starting location to a single destination. A user may need to visit multiple locations, requiring the planner to determine a visiting order that reduces total travel distance while avoiding static and dynamic obstacles. For example, a resident in an assisted-living facility may need to travel from a bedroom to a medication station, dining area, and outdoor space before returning to the room, while another resident may need to visit several service counters in a cafeteria before proceeding to a checkout point and seating area~\cite{sinyukov2019wheelchairs, desmond2013development}. This is particularly relevant in structured indoor environments, where the static layout is known in advance while the workspace remains dynamically occupied by humans and other wheelchair users. Such scenarios require an autonomous wheelchair to combine efficient multi-destination path-planning with continuous adaptation to surrounding static and dynamic obstacles, as illustrated in Fig.~\ref{fig:teaser}.
%when the static map of the environment is known and the user moves in a controlled environment that is full of humans or other wheelchair users. 

%Paragraph2: how this problem is solved in literature?
%What is the distinction between global and local planning in the literature, and what motivates hybrid/hierarchical architectures?
Path planning in complex and dynamic environments requires a robot not only to determine an efficient route toward its destination, but also to adapt that route in response to dynamic obstacles to avoid collisions~\cite{lu2025global}. Existing path-planning approaches for autonomous wheelchairs and mobile robots are therefore often organized into global and local planning layers. Global planners use a static map of the environment to generate long-range routes between an initial position and a destination, typically by determining an optimal or near-optimal path between the start and goal. This planning process is generally performed offline, assuming that the static structure of the environment is known in advance~\cite{wang2020mobile,li2024path, debnath2024integrated}. Common approaches include graph-search methods such as A*~\cite{jian2021global,li2024path} and Dijkstra’s algorithm~\cite{jung2020path} due to their simplicity and effectiveness in shortest path problems, as well as evolutionary computation techniques~\cite{lu2025global, debnath2024integrated, guo2020global}. Local planners typically generate short-range motion plans to avoid nearby dynamic obstacles and are often executed online as the environment changes~\cite{wang2020mobile,li2024path, debnath2024integrated}. Examples include Artifical PF~\cite{guan2025improved,rasekhipour2016potential, khatib1985real, khatib1986real}, RRT and their variants~\cite{lavalle2001randomized,karaman2011sampling,wang2020neural,guan2025improved}, learning-based methods~\cite{wang2020mobile,wang2020neural, schmuck2025sang}, the Dynamic Window Approach (DWA)~\cite{lu2025global}, and Model Predictive Control (MPC)~\cite{li2024path,rasekhipour2016potential}. Given the complementary roles of global and local planning, hierarchical and hybrid architectures have been widely investigated to combine long-range route generation with short-range collision avoidance algorithms~\cite{wang2020mobile,jung2020path,jian2021global,lu2025global,li2024path}. For PWs operating in structured environments such as hospitals, nursing homes, and assisted-living facilities, integrating global and local path planning is particularly important. Efficient travel among multiple destinations should be achieved while continuously avoiding collisions and maintaining user safety in the presence of pedestrians and other wheelchair users~\cite{jung2020path}.

When a robot must visit multiple task locations, however, the problem
becomes significantly more complex. A common approach to addressing multi-goal planning is to determine the optimal visiting order of goals, which can
be formulated as a TSP~\cite{guo2020global, faigl2019fast, janovs2021multi, debnath2024integrated, bhat2026parallel}. Graph-search methods such as A* or Dijkstra's algorithm can also be used as global path planners (GPPs) for shortest-path problems~\cite{jung2020path, sinyukov2014multi}. When the number of waypoints is relatively
small, an A*-based multi-goal formulation can produce the same global path as the TSP-based approach. However, its computational cost increases significantly as the number of waypoints
grows, making it impractical for large waypoint sets~\cite{huang2025self}. The TSP is a classical optimization problem that seeks an optimal path to visit all targets exactly once~\cite{yang2015improved, xie2018integrated}. By computing an efficient global visiting sequence, the robot can minimize the overall travel
distance required to complete multiple tasks. However, when dynamic or randomly
moving obstacles are present in the environment, a TSP-based GPP alone is insufficient for handling dynamic obstacles and therefore needs to be complemented by a local planning mechanism capable of real-time obstacle avoidance strategies. Despite their strengths, planners such as A* and RRT can face computational challenges in highly dynamic and complex environments. As the size and complexity of the environment increase, the computational
burden of these algorithms may also increase, which can hinder reliable real-time operation~\cite{guan2025improved, huang2025self, wang2020neural, zhang2018path}. For these reasons, a practical strategy is to combine global path-planning methods such as TSP with local motion planners such as RRT or A*~\cite{bhat2026lazy}.
In such hybrid frameworks, TSP determines the global ordering of tasks, while
local planners can handle short-range path generation and obstacle avoidance algorithms. However, an important challenge in these approaches is how to effectively transfer global route information generated by the GPP to heuristic local planners (HLPs) operating in dynamic environments~\cite{jian2021global}.

%Paragraph4: Previous Work and Limitations
%Paragraph5: Contributions
%Paragraph6: Paper Organization

Although prior studies have demonstrated effective hybrid planning strategies and dynamic obstacle avoidance, many approaches are evaluated primarily in simulation using simplified robot models and relatively low-density dynamic environments, or purely static environments~\cite{lu2025global, wang2020mobile, li2024path, debnath2024integrated, jian2021global, jung2020path, guo2020global, guan2025improved, rasekhipour2016potential, wang2020neural, janovs2021multi, huang2025self, xie2018integrated, zhang2018path, sinyukov2014multi}. Fewer studies investigate the complete path-planning pipeline on a physical PW operating in a dynamic environment~\cite{loke2026minimal, debnath2024integrated, guo2020global}. Moreover, maintaining consistency between the global and local planning layers remains an important challenge, since local obstacle-avoidance maneuvers can cause the robot to deviate from the global path. While several studies have explored global-local coupling strategies~\cite{jian2021global, wang2020mobile}, to the best of our knowledge, our previous work~\cite{leblebicioglu2026hybrid} was the first to directly transfer information from a global multi-goal path to local planners through PF formulation using dedicated guidance forces. Beyond global-local coordination, physical feasibility is another key consideration. Vehicle kinematic equations have been incorporated into some path-planning frameworks~\cite{li2024path, jung2020path, guan2025improved}; however, curvature constraints remain essential for generating realistic and executable paths~\cite{jian2021global, guo2020global, janovs2021multi}. The physical size and shape of the wheelchair must also be considered to ensure collision-free motion~\cite{sinyukov2014augmenting, greenberg2025effects}. 

Building on these observations,~\cite{leblebicioglu2026hybrid} introduced a hybrid multi-goal navigation framework integrating TSP-based global planning, PF guidance, and HLPs, resulting in two hybrid approaches: TSP–PF–RRT and TSP–PF–A*. A central contribution of that work was the use of the PF formulation as an interface between the GPP and HLPs. Specifically, two additional path-guidance forces, a pulling force ($\mathbf{F}_{pull}$) and a tangential force ($\mathbf{F}_{tangent}$), were introduced to transfer global path information generated by the TSP-based GPP to the local RRT or A* planner. $\mathbf{F}_{pull}$ attracts the wheelchair toward the global path when it is disturbed by obstacle-avoidance maneuvers, whereas $\mathbf{F}_{tangent}$ provides directional guidance along the path. Together with the attractive and repulsive forces, the resulting total force defines the local guidance direction used by RRT or A*, allowing the wheelchair to avoid static and dynamic obstacles while remaining guided by the globally optimized multi-goal route. The framework was evaluated primarily through Monte Carlo simulations in crowded and extra-crowded dynamic environments. However, the study remained simulation-based and did not evaluate the complete framework on a physical PW. Furthermore, the earlier formulation relied on a simplified robot representation and did not explicitly investigate shape-aware collision checking for path feasibility.

%buraya websitesini referans verelim. 
The present work extends~\cite{leblebicioglu2026hybrid} by introducing a higher-fidelity simulation framework and extensive experimental validation on a physical PW. The main extensions beyond the previous work are summarized as follows: 

\begin{itemize}
    \item \textbf{Higher-fidelity wheelchair modeling:}
    The simulation framework is redesigned to better capture the physical geometry of the wheelchair. In the previous study, the wheelchair was represented using a point-based model, whereas in the present work it is modeled with an elliptical geometry that accounts for its physical dimensions in repulsive-force calculations involving both static and dynamic obstacles~\cite{jian2021global, greenberg2025effects}. 

    \item \textbf{Enhanced local planning:}
    The A* planner employs an extended neighborhood with direction-dependent motion costs, while the RRT planner incorporates environment-informed sampling and a two-step tree-extension strategy to improve local exploration in dynamic environments. To the best of our knowledge, the integration of these A*- and RRT-specific enhancements into a PF-guided local-planning framework for autonomous wheelchair navigation has not previously been reported (see Section~\ref{sec4}). 
    
    \item \textbf{Real-world experimental validation:}
    The proposed TSP--PF--RRT and TSP--PF--A* frameworks are implemented and evaluated on a physical wheelchair through two complementary experimental settings. MoCap-based experiments with human pedestrians are conducted to characterize the planning behavior using accurate ground-truth measurements of both pedestrian and wheelchair motion, while experiments using onboard LiDAR and RGB-D cameras are conducted to evaluate the complete planning framework under onboard perception. These two experimental settings therefore provide complementary evaluations under controlled ground-truth measurements and realistic onboard sensor-based planning. 

    % \item \textbf{Simulation-to-real accessibility scenario:}
    % A van-ramp scenario is introduced as a simulation-to-real case study representative of an assistive mobility task. The scenario is evaluated both in simulation and on the physical wheelchair, with a moving pedestrian intentionally crossing the wheelchair trajectory to examine the response of the planning framework under a realistic dynamic interaction.
\end{itemize}

%websitesi/youtube video referans. 
The remainder of this study is organized as follows: Section 2 introduces the global path-planning and transformation of global path information to local heuristics with PF method. Section 3 describes the dynamic obstacle model and formulates robot–obstacle separation for both point-based and elliptical wheelchair representations. Section 4 details improved HLPs: enhanced RRT and modified A*. Section 5 outlines the simulation framework and evaluates both point-based and elliptical wheelchair models under normal, crowded, and extra-crowded dynamic environments. Section 6 explains the physical wheelchair platform and control structure. Section 7 reports experiments using MoCap and onboard LiDAR/RGB-D sensing. Finally, Section 8 summarizes the main findings, discusses limitations, while Section 9 outlines the future work, and concludes the study.

\section{Global Path Planning with TSP}\label{sec2}
\begin{figure}[h]
\centering
\includegraphics[width=0.65\textwidth]{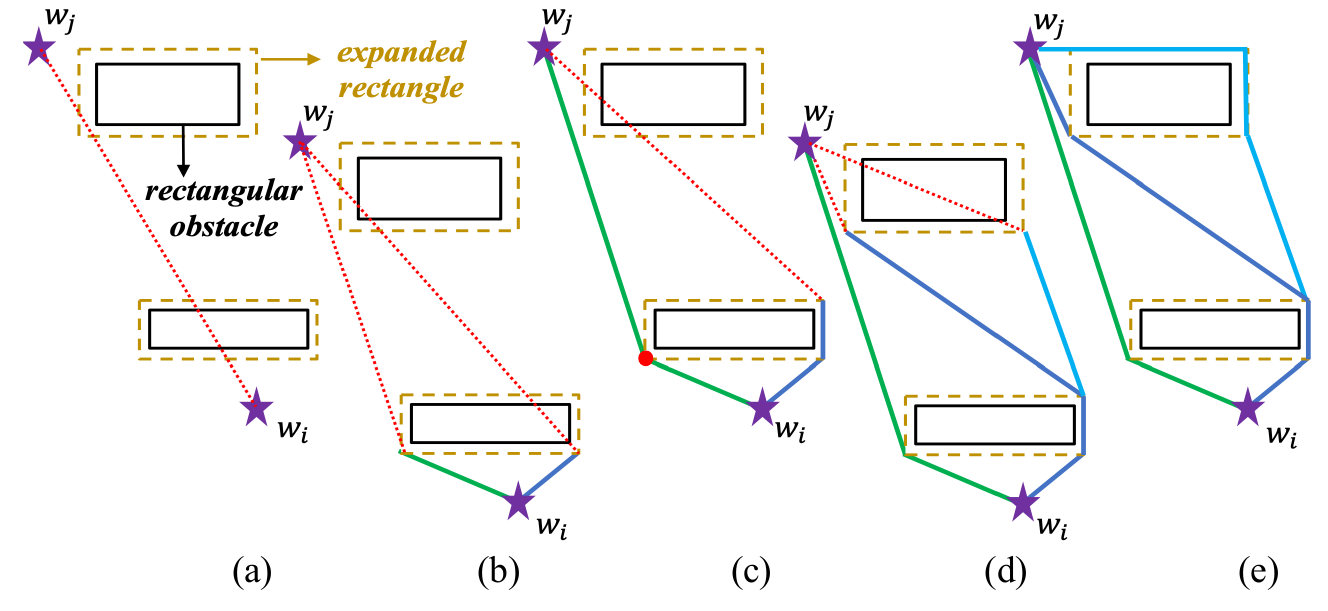}
\caption{Construction of all feasible connections between waypoints ($w_i$--$w_j$) before solving the TSP: (a) direct connection, (b) routing around the corners of the expanded rectangle of the first intersected obstacle, (c) completed (green) and continuing (blue) candidate paths, (d) repetition of the routing, and (e) all feasible collision-free paths. The TSP-based global planner computes a path around these expanded rectangles, thereby avoiding the static obstacles.}
\label{fig:TSP}
\end{figure}

The path-planning environment contains heterogeneous obstacles, including rectangular static obstacles and elliptical dynamic obstacles representing pedestrians and other wheelchairs in settings such as nursing homes. The dynamic obstacles are assigned different sizes and velocities to represent variations in physical dimensions and motion. Only static rectangular obstacles are considered during TSP-based global path planning; dynamic obstacles are handled by the local planner. Determining a path that visits each waypoint exactly once while minimizing the total travel distance corresponds to the classical TSP. Due to the availability of efficient solution methods, TSP is employed to compute the global path,~\cite{faigl2019fast}. To ensure safe navigation, each rectangular obstacle is expanded by a safety margin of 1~m (safety margin is different in hardware experiments to account for the dimensions of the test space) in all directions, forming an expanded rectangle that prevents the robot from entering the obstacle region (illustrated as dashed yellow rectangles in Fig.~\ref{fig:TSP}). 

% \begin{figure}[h]
% \centering
% \includegraphics[width=1.0\textwidth]{figures/TSP.pdf}
% \caption{Construction of all feasible connections between the start, waypoints ($w_i$-$w_j$) and the
% goal before solving the TSP. The TSP-based global planner computes a path around these expanded rectangles, thereby avoiding the static obstacles.}\label{fig:TSP}
% \end{figure}

The objective of the global planner is to construct all feasible geometric connections between the start position, waypoints, and the goal while avoiding the static obstacles (illustrated as black rectangles in Fig.~\ref{fig:TSP}). Among these connections, the shortest path that visits all waypoints is determined using particle swarm optimization (PSO). The global path is recomputed every 10~s from the current wheelchair position through the remaining waypoints, to account for possible changes in the environment.

The GPP is constructed as follows:
\begin{enumerate}
    \item Let
    $V=\{q_0,w_1,w_2,\ldots,w_N,q_{\mathrm{goal}}\}$
    denote the set of nodes consisting of the start position,
    waypoints, and goal. For each pair of nodes in $V$, a direct
    geometric connection is first considered by drawing a
    straight line between them (i.e., the red dotted line between
    stations $w_i$ and $w_j$ in Fig.~\ref{fig:TSP}a). If the line does not
    intersect any rectangular obstacle, the connection is
    accepted.

    \item If the straight line intersects a rectangular obstacle,
    the algorithm identifies the first intersecting rectangle
    along the connection and considers paths through the corners
    of its expanded rectangle. Only collision-free connections
    to these corners are retained (green and blue lines in Fig.~\ref{fig:TSP}b).

    \item During this process, corners already visited along
    the current candidate path are not revisited, avoiding
    redundant paths and cycles (i.e., in Fig.~\ref{fig:TSP}c, the blue path does not return to the corner marked with the red dot, since this corner has already been visited).

    \item At each step, the algorithm checks whether a direct
    collision-free connection from the current node to the
    destination node exists. If so, the candidate path is
    completed (green path in Fig.~\ref{fig:TSP}c). Otherwise, the search
    continues through feasible corners of the expanded
    rectangles (blue paths in Figs.~\ref{fig:TSP} c and d), until all feasible collision-free paths between the two nodes are obtained (Fig.~\ref{fig:TSP}e).

    \item Once all feasible geometric connections between the
    start position, waypoints, and goal have been constructed,
    a topological graph is formed. PSO is then applied to solve
    the resulting TSP. The global path obtained using PSO for
    an 11-waypoint problem is shown on the left in Fig.~\ref{fig:PF}.
\end{enumerate}

\subsection{Transformation of Global Path Information to Local Heuristics with PF Method}\label{sec2.1}
PF is used together with HLPs: RRT or A* for local path planning of the robot (wheelchair). The PF provides a directional guidance force, while RRT or A* ensure collision-free motion by locally planning feasible paths. In the PF framework, the robot is attracted toward the goal with lower potential and repelled from obstacles with higher potential through virtual attractive ($\mathbf{F}_{\text{attractive}}$) and repulsive forces ($\mathbf{F}_{\text{repulsive}}$) generated by the potential field,~\cite{guan2025improved}. This behavior can be explained through a physical analogy: the robot acts as a positively charged particle, repelled by positively charged obstacles and attracted to a negatively charged goal. The formulation for calculating $\mathbf{F}_{\text{attractive}}$ towards stations is given in Eq.~\ref{eq:attractive}, where $k_{att}$ is the coefficient of attraction, $\mathbf{q} = [q_x, q_y]$ is the current position of the robot, and $\mathbf{q}_{goal} = [q_{goalx}, q_{goaly}]$ is the position of waypoints, expanded rectangle corners and the end goal. The parameter $\epsilon$ is a small positive constant, with $\boldsymbol{\epsilon} = [\epsilon,\, \epsilon]^\top$, that prevents $\mathbf{F}_{\text{attractive}}$ from becoming zero or unbounded.

% \begin{equation}
% \label{eq:attractive}
% \mathbf{F}_{att} = 
% k_{att}\frac{\mathbf{q}_{goal}-\mathbf{q}}
% {\|\mathbf{q}_{goal}-\mathbf{q}\|}
% \end{equation}

\begin{equation}
\label{eq:attractive}
\mathbf{F}_{att} =
k_{att}\frac{(\mathbf{q}_{goal}-\mathbf{q})+\boldsymbol{\epsilon}}
{\|\mathbf{q}_{goal}-\mathbf{q}\| + \epsilon}
\end{equation}

Static obstacles are modeled as rectangles. Each rectangle is inflated by a margin $\beta$. When the robot is modeled as a point, $\beta$ acts as a safety margin, so that the repulsive force starts at a certain distance from the obstacle itself. When the robot is modeled as an ellipse, $\beta$ is set to its semi-major axis to take into account the robot's body. In both cases, $\beta$ is smaller than the buffer used to expand the obstacles in the global planner; hence, the corners of the global path remain outside this boundary and the robot can reach them while keeping a safe distance from the obstacles.

For each rectangle, a signed distance $d_i$ is computed between the robot
position $\mathbf{q}$ and the inflated rectangle. Its magnitude is the
distance from the robot to the nearest point $\mathbf{p}_i^{*}$ on the
boundary ($|d_i| = \|\mathbf{q} - \mathbf{p}_i^{*}\|$), and its sign indicates whether the robot is outside ($d_i > 0$) or
inside ($d_i \le 0$) the inflated rectangle. In both cases, the direction
of the repulsive force is
$\hat{\mathbf{n}}_i = (\mathbf{q} - \mathbf{c}_i) / \|\mathbf{q} - \mathbf{c}_i\|$, where $\mathbf{c}_i$ is the center of the rectangle; hence, the robot is pushed away from the center
of the static obstacle. The repulsive force applied by a rectangle is then
computed as in Eq.~\ref{eq:Frect}, where $k_{rep}$ is the repulsion
coefficient, $d_0^{rect}$ is the predefined influence distance, and $F_{in}$ is a
constant force magnitude applied when the robot is inside the inflated
rectangle to strongly repel the robot. The total repulsive force generated by all rectangular obstacles is
obtained by summing the individual forces,
$\mathbf{F}_{rep}^{rect} = \sum_i \mathbf{F}_{rect}^{(i)}$~\cite{guan2025improved}.

\begin{equation}
\label{eq:Frect}
\mathbf{F}_{rect}^{(i)} =
\begin{cases}
F_{in}\, \hat{\mathbf{n}}_i, & d_i \le 0,\\[4pt]
k_{rep} \left(\dfrac{1}{d_i} - \dfrac{1}{d_0^{rect}}\right) \dfrac{1}{d_i^{2}}\, \hat{\mathbf{n}}_i,
  & 0 < d_i < d_0^{rect},\\[4pt]
\mathbf{0}, & d_i \ge d_0^{rect}.
\end{cases}
\end{equation}

%To simplify the distance computation, each ellipse is mapped into a unit circle centered at the origin. 
Dynamic obstacles are modeled as ellipses with centers
$\mathbf{o}_i$. For each obstacle, a normalized distance $d_E$ is computed between the robot
position $\mathbf{q}$ and the ellipse, where $d_E = 0$ corresponds to the
ellipse boundary (see Section~\ref{sec3.1}). When $d_E < d_0^{ell}$ ($d_0^{ell}$ is the predefined influence distance for elliptical obstacles), the
obstacle generates a repulsive force acting on the robot
(Eq.~\ref{eq:Fellipse}). The magnitude of this force is defined as a piecewise constant value. In the outer part of the influence region, the force magnitude is kept small to allow the robot to smoothly maneuver around the obstacle ($k_{low}$). When the robot approaches a safety threshold $d_{\text{safe}}$ close to the obstacle boundary, the force magnitude switches to a significantly larger value ($k_{high}$) in order to strongly repel the robot and prevent collisions. The direction of the repulsive force is defined along the vector from the obstacle center to the robot position and a tangential component is added to this direction in Section~\ref{4.3}. The repulsive force generated by all elliptical obstacles is obtained by summing the individual forces $\mathbf{F}_{rep}^{ellipse} = \sum_i \mathbf{F}_{ell}^{(i)}$. The total repulsive force acting on the robot is $\mathbf{F}_{\textit{repulsive}} = \mathbf{F}_{\textit{rep}}^{\textit{rect}} + \mathbf{F}_{\textit{rep}}^{\textit{ellipse}}$.

\begin{equation}
\mathbf{F}_{ell}^{(i)} =
\begin{cases}
k_{low} \dfrac{\mathbf{q}-\mathbf{o}_{i}}{\|\mathbf{q}-\mathbf{o}_{i}\|}, & d_{safe} < d_E < d_0^{ell} \\[6pt]
k_{high} \dfrac{\mathbf{q}-\mathbf{o}_{i}}{\|\mathbf{q}-\mathbf{o}_{i}\|}, & d_E \le d_{safe} \\[6pt]
0, & d_E \ge d_0^{ell}
\end{cases}
\label{eq:Fellipse}
\end{equation}

\begin{figure}[h]
\centering
\includegraphics[width=0.75\textwidth]{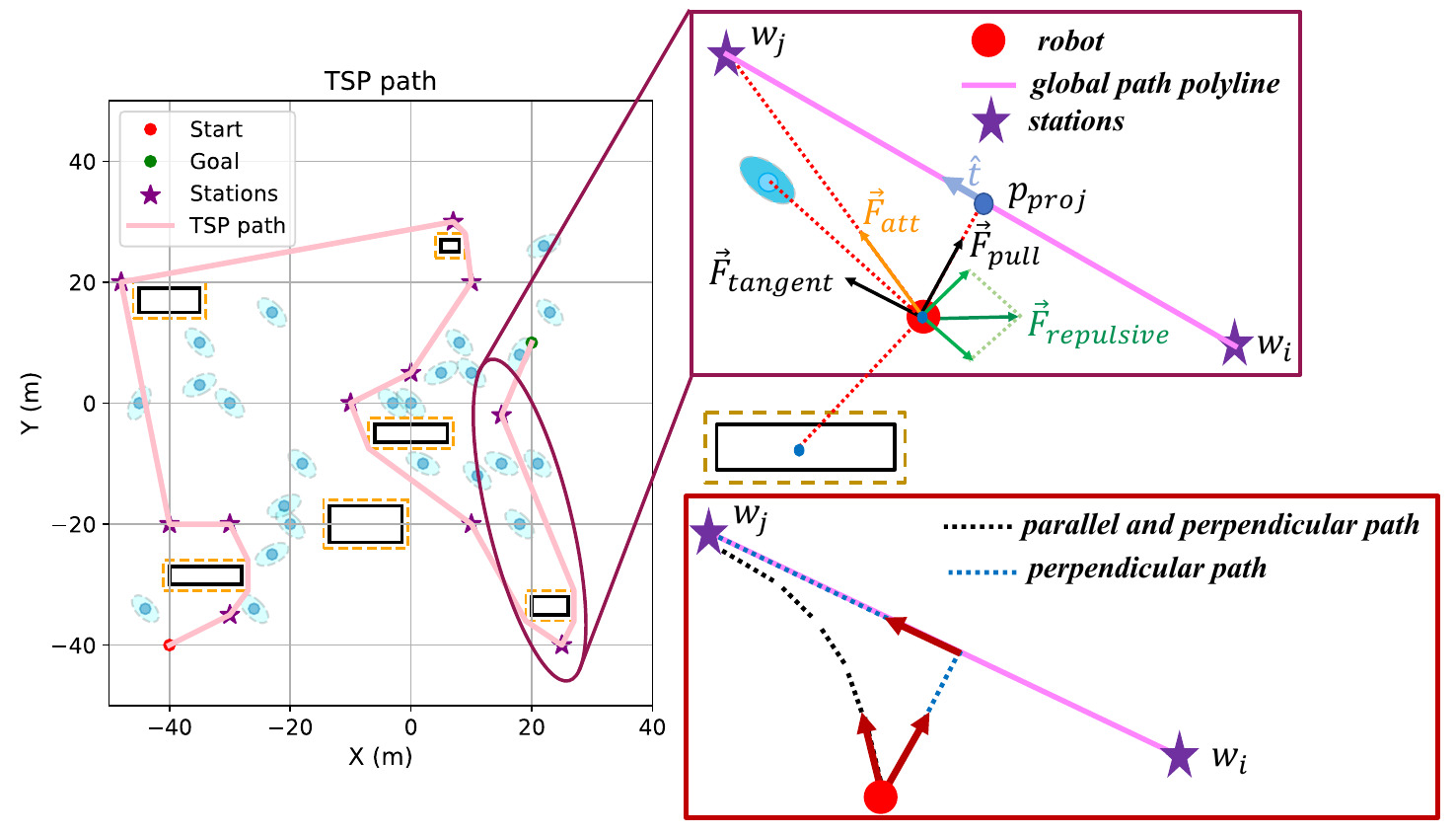}
\caption{Left: Global optimal path for an 11-waypoint problem obtained using PSO,
starting from the initial position, visiting each station once, and finally
reaching the goal. Right: PF components used for navigation between stations
$w_i$ and $w_j$: $F_{\text{tangent}}$ and $F_{\text{pull}}$ guide the robot
along the global path, $F_{\text{attractive}}$ pulls the robot toward the next
station, and $F_{\text{repulsive}}$ represents repulsive forces generated by
rectangular and elliptical obstacles. With the combined effect of
$F_{\text{tangent}}$ and $F_{\text{pull}}$, the robot follows a
hypotenuse-like trajectory to rejoin the global path through a shorter route.}\label{fig:PF}
\end{figure}

There are two additional forces that are associated with the global path. These are the tangent force along the global path, $\mathbf{F}_{\text{tangent}}$, and a pulling force, $\mathbf{F}_{\text{pull}}$, which ensure that the robot follows the global path even when it is disturbed by the repulsive forces generated by obstacles ($\mathbf{F}_{\text{repulsive}}$). In Eq.~\ref{eq:Fglobal}, $\mathbf{p}_{\text{proj}}$ denotes the closest point to the robot position $\mathbf{q}$ on the global path polyline, and $\hat{\mathbf{t}}$ denotes the unit tangent vector of the corresponding path segment, $k_{pull}$ and $k_{tangent}$ are attractive force coefficients towards the global path. The pulling gain \(k_{\text{pull}}\) is constant, whereas the tangent gain \(k_{\text{tan}}\) depends on the perpendicular distance \(d_{\perp}=\|\mathbf{p}_{\text{proj}}-\mathbf{q}\|\) from the robot to the global path polyline and decreases as the robot approaches the path. The total force acting on the robot is $\mathbf{F}_{\text{total}} =
\mathbf{F}_{\text{attractive}} +
\mathbf{F}_{\text{repulsive}} +
\mathbf{F}_{\text{tangent}} +
\mathbf{F}_{\text{pull}}$, shown in the top-right side of Fig.~\ref{fig:PF}. When the robot deviates from the global path, the combined effect of $F_{\text{tangent}}$ and $F_{\text{pull}}$ drives it along an approximately hypotenuse-like trajectory, rather than first moving perpendicularly back to
the path and then following it. The robot therefore rejoins the global path,
and approaches the next waypoint, through a shorter route, as illustrated in
the bottom-left side of Fig.~\ref{fig:PF}.

\begin{equation}
\label{eq:Fglobal}
\mathbf{F}_{\text{pull}} = k_{\text{pull}} (\mathbf{p}_{\text{proj}} - \mathbf{q}), \qquad
\mathbf{F}_{\text{tangent}} = k_{\text{tan}}(d_{\perp})\hat{\mathbf{t}}
\end{equation}

\section{Dynamic Obstacle Modeling}\label{sec3}
% stochastic maneuver,
% noisy center,
% circle to ellipse approximation, radius of the ellipse boundary
% in the direction of the robot.

Dynamic obstacles are modeled as ellipses of heterogeneous size. Each obstacle $i$ is assigned a size coefficient $s_i$, which scales the base semi-major and semi-minor axes $a_0 = 0.5$\,m and $b_0 = 0.4$\,m to obtain obstacle-specific base dimensions $a_{\text{base},i} = s_i a_0$ and $b_{\text{base},i} = s_i b_0$. The size coefficients are assigned per obstacle within the range $[0.8,\, 2.0]$. The dimensions and orientation of each ellipse depend on its velocity $\mathbf{v}_i = [v_{x,i}, v_{y,i}]^\top$. The semi-major axis $a_i$ is aligned with the direction of velocity, while the semi-minor axis $b_i$ is perpendicular to it, so that faster obstacles occupy a larger region along their direction of travel. Such non-circular, motion-dependent regions are consistent with recent proxemics studies~\cite{camara2025proxemic}. The axis lengths and the orientation are defined in Eq.~\ref{eq:a_b}, where $\|\mathbf{v_{i}}\|$ is the velocity magnitude, $\alpha = 0.2$ and $\beta = 0.1$ are scaling coefficients, $a_{\max} = 3$\,m and $b_{\max} = 1.5$\,m are the upper bounds on the axes, and $\theta_i$ is the angle between the global $x$-axis and the major axis of the ellipse. The velocity magnitude of each obstacle is bounded as $v_{\min} \leq \|\mathbf{v}_i\| \leq v_{\max}$, with $v_{\min} = 5.0$ and $v_{\max} = 24.5$.

\begin{equation}
\label{eq:a_b}
\begin{aligned}
a_i &= \min\left(a_{\text{base},i} + \alpha \|\mathbf{v}_i\|,\; a_{\max}\right), \\
b_i &= \min\left(b_{\text{base},i} + \beta \|\mathbf{v}_i\|,\; b_{\max}\right), \\
\theta_i &= \operatorname{atan2}\left(v_{y,i},\, v_{x,i}\right),
\end{aligned}
\end{equation}

Stochastic maneuvering behavior is also introduced for the dynamic obstacles. Specifically, an obstacle may exhibit small speed fluctuations (i.e., slight increases or decreases in speed), large random maneuvers, or small random drifts in its direction of velocity. To simulate sensing uncertainty, Gaussian noise with zero mean and a standard deviation of $\sigma = 0.1~\mathrm{m}$ is added to the elliptical obstacle positions. The perceived obstacle positions are modeled in Eq.~\ref{eq:obstacle_noise}, where $\mathbf{p}_{\mathrm{obs}}^{\mathrm{true}}$  and $\mathbf{p}_{\mathrm{obs}}^{\mathrm{noisy}}$ denote the true and noisy obstacle positions, respectively.

\begin{equation}
\label{eq:obstacle_noise}
\mathbf{p}_{\mathrm{obs}}^{\mathrm{noisy}} = \mathbf{p}_{\mathrm{obs}}^{\mathrm{true}} + \boldsymbol{\epsilon}, 
\qquad 
\boldsymbol{\epsilon} \sim \mathcal{N}\!\left(\mathbf{0},\, \sigma^2\right)
\end{equation}

\subsection{Robot--Obstacle Separation for Point and Elliptical Wheelchair Models}\label{sec3.1}

\begin{figure}[h]
\centering
\includegraphics[width=0.5\textwidth]{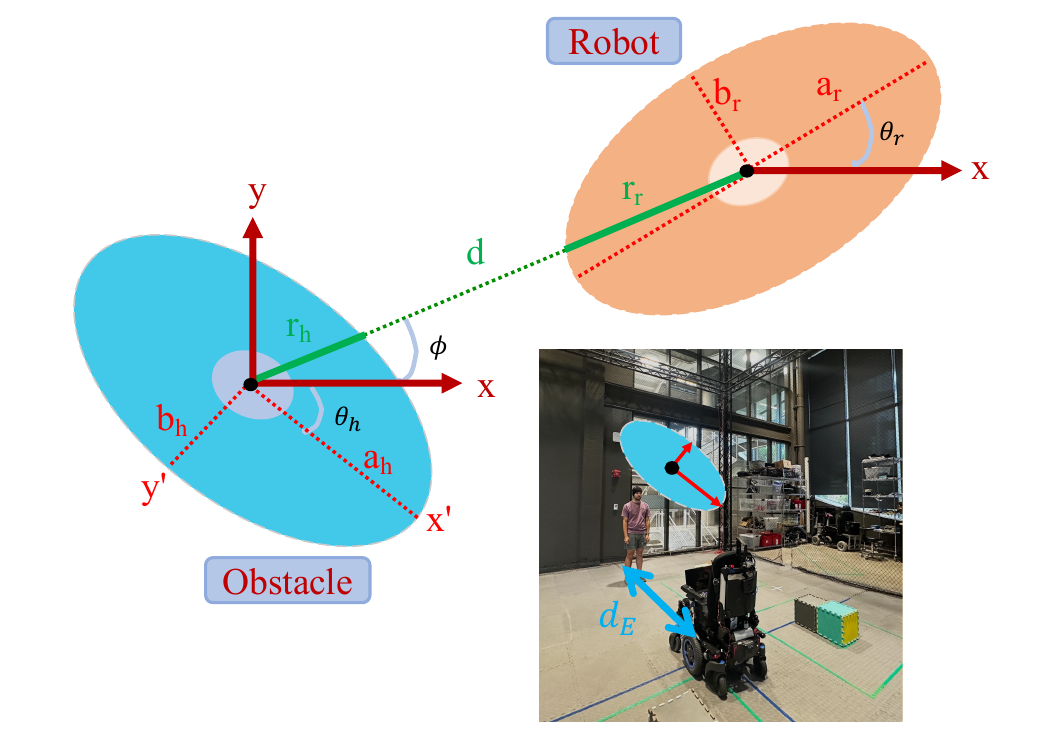}
\caption{Obstacle--robot ellipse geometry showing $d$, $\phi$, radial extents $r_h$
and $r_r$ along the line of centers, and orientations $\theta_h$ and
$\theta_r$.}\label{fig:ellipse_geometry}
\end{figure}

Let the centers of the robot and the obstacle be denoted by
$\bar{\mathbf{q}}=[q_x,\;q_y]^\top$ and
$\bar{\mathbf{o}}=[o_x,\;o_y]^\top$, respectively, in the world coordinate
frame. The obstacle is modeled as an ellipse whose major axis is rotated by
an angle $\theta_h$ relative to the world $x$-axis, where $\theta_h$ is
computed from the direction of the obstacle's velocity and therefore changes as the obstacle moves (an example configuration is shown in Fig.~\ref{fig:ellipse_geometry}). To express the robot center in the obstacle frame, its position relative to
the obstacle center, $[q_x-o_x,\;q_y-o_y]^\top$, is rotated by
$-\theta_h$ about the $z$-axis, which is perpendicular to the $x$-$y$ plane.
The resulting coordinates $(x'_h,y'_h)$ are given in
Eq.~\eqref{eq:obstacle_frame_coordinates}.

\begin{equation}
    \begin{bmatrix}
        x'_h\\
        y'_h
    \end{bmatrix}
    =
    \underbrace{
    \begin{bmatrix}
        \cos\theta_h & \sin\theta_h\\
        -\sin\theta_h & \cos\theta_h
    \end{bmatrix}}_{R(-\theta_h)}
    \begin{bmatrix}
        q_x-o_x\\
        q_y-o_y
    \end{bmatrix}.
    \label{eq:obstacle_frame_coordinates}
\end{equation}

Let $a_h$ and $b_h$ denote the obstacle ellipse's semi-axis lengths along its
local $x'$- and $y'$-axes, respectively. In the obstacle frame, an arbitrary
point $(x,y)$ on the obstacle boundary satisfies the ellipse equation given
in Eq.~\eqref{eq:obstacle_ellipse}. Evaluating the ellipse expression at the robot center gives the normalized
distance measure $\rho_E$, as defined in Eq.~\eqref{eq:rho_ellipse}. As given in Eq.~\eqref{eq:rho_ellipse}, $\rho_E=1$ when the robot center lies
on the ellipse boundary, $\rho_E<1$ when it lies inside, and $\rho_E>1$ when
it lies outside. The point-robot distance measure is defined as
$d_E^{\mathrm{point}}=\rho_E-1$. Thus,
$d_E^{\mathrm{point}}>0$ when the robot center lies outside the ellipse,
$d_E^{\mathrm{point}}=0$ when it lies on the boundary, and
$d_E^{\mathrm{point}}<0$ when it lies inside. This is a dimensionless
normalized distance.

\begin{equation}
    \left(\frac{x}{a_h}\right)^2
    +
    \left(\frac{y}{b_h}\right)^2
    =1
    \label{eq:obstacle_ellipse}
\end{equation}

\begin{equation}
    \rho_E =
    \sqrt{
        \left(\frac{x'_h}{a_h}\right)^2
        +
        \left(\frac{y'_h}{b_h}\right)^2
    }
    \label{eq:rho_ellipse}
\end{equation}

Alternatively, the same distance measure can be obtained by mapping the
obstacle ellipse to a unit circle through coordinate scaling. Define the
normalized coordinates as $\tilde{x}_h=x'_h/a_h$ and
$\tilde{y}_h=y'_h/b_h$. Under this coordinate transformation, the ellipse
boundary in Eq.~\eqref{eq:obstacle_ellipse} becomes the unit circle
$\tilde{x}_h^2+\tilde{y}_h^2=1$. The robot center's radial coordinate in the
normalized frame is $\rho_E=\sqrt{\tilde{x}_h^2+\tilde{y}_h^2}$. Subtracting
the unit boundary radius gives
$d_E^{\mathrm{point}}=\rho_E-1
=\sqrt{\tilde{x}_h^2+\tilde{y}_h^2}-1$, which is the same point-robot
distance measure defined above. 

An equivalent expression for $\rho_E$ can be obtained from the distance and direction between the two centers. The center-to-center distance $d$ and
world-frame angle $\phi$ are given in Eq.~\eqref{eq:center_distance_angle}.

\begin{equation}
    d =
    \sqrt{(q_x-o_x)^2+(q_y-o_y)^2},
    \qquad
    \phi =
    \operatorname{atan2}(q_y-o_y,\;q_x-o_x).
    \label{eq:center_distance_angle}
\end{equation}

Here, $\phi$ is measured from the world $x$-axis. Since the obstacle's major
axis is oriented by $\theta_h$, the direction from the obstacle center to
the robot center, expressed in the obstacle frame, is
$\psi_h=\phi-\theta_h$. Consequently, the obstacle-frame coordinates given in
Eq.~\eqref{eq:obstacle_frame_coordinates} can also be expressed as
in Eq.~\eqref{eq:obstacle_frame_polar_coordinates}.

\begin{equation}
    x'_h=d\cos\psi_h,
    \qquad
    y'_h=d\sin\psi_h.
    \label{eq:obstacle_frame_polar_coordinates}
\end{equation}

Substituting Eq.~\eqref{eq:obstacle_frame_polar_coordinates} into
Eq.~\eqref{eq:rho_ellipse} gives the equivalent expression for $\rho_E$
defined in Eq.~\eqref{eq:rho_angle_form}.

\begin{equation}
    \rho_E =
    d\sqrt{
        \left(\frac{\cos\psi_h}{a_h}\right)^2
        +
        \left(\frac{\sin\psi_h}{b_h}\right)^2
    }.
    \label{eq:rho_angle_form}
\end{equation}

The obstacle's boundary radius along this direction, $r_h$, is given in
Eq.~\eqref{eq:obstacle_radial_radius}. Comparing Eqs.~\eqref{eq:rho_angle_form} and
\eqref{eq:obstacle_radial_radius} gives $\rho_E=d/r_h$. Combining this result with the point-robot distance definition gives
$d_E^{\mathrm{point}}=d/r_h-1$.

\begin{equation}
    r_h =
    \frac{1}{
        \sqrt{
            \left(\frac{\cos\psi_h}{a_h}\right)^2
            +
            \left(\frac{\sin\psi_h}{b_h}\right)^2
        }
    }.
    \label{eq:obstacle_radial_radius}
\end{equation}

The formulation can be extended to a robot modeled as an ellipse. Let $a_r$
and $b_r$ denote the robot's semi-axis lengths, and let $\theta_r$ denote
the orientation of its major axis relative to the world $x$-axis. The angle
$\phi$ points from the obstacle center toward the robot center; therefore,
the direction from the robot center toward the obstacle is $\phi+\pi$.
Measured relative to the robot's major axis, this direction is
$\psi_r=\phi+\pi-\theta_r$. 
The robot's boundary radius along the line joining the two centers, $r_r$,
is given in Eq.~\eqref{eq:robot_radial_radius}.

\begin{equation}
    r_r =
    \frac{1}{
        \sqrt{
            \left(\frac{\cos\psi_r}{a_r}\right)^2
            +
            \left(\frac{\sin\psi_r}{b_r}\right)^2
        }
    }.
    \label{eq:robot_radial_radius}
\end{equation}

Using the sum of the obstacle and robot radii along the centerline, the
two-body distance measure is defined in Eq.~\eqref{eq:pair_level_set}.

\begin{equation}
    d_E^{\mathrm{pair}}
    =
    \frac{d}{r_h+r_r}-1.
    \label{eq:pair_level_set}
\end{equation}

Under this centerline radial measure,
$d_E^{\mathrm{pair}}=0$ corresponds to contact along the line joining the
centers, and $d_E^{\mathrm{pair}}<0$ indicates overlap along that line.
When $d_E^{\mathrm{pair}}>0$, a gap exists along the centerline. For a point robot, $r_r=0$. In this case, the two-body distance measure in
Eq.~\eqref{eq:pair_level_set} reduces to the point-robot distance measure:
$d_E^{\mathrm{pair}}=d/r_h-1=d_E^{\mathrm{point}}$.

\section{Local Path Planning with enhanced RRT and modified A*}\label{sec4}

In this section, two HLPs, PF-RRT and PF-A*, are introduced. The PF method is used together with RRT or A* for obstacle avoidance, forming the local layer of the overall motion planning. Local path planning, obstacle avoidance, and tracking of the optimal global path are performed by the HLPs, while the PF technique conveys the global path information from the GPP (i.e., TSP), to the local planner. Compared to~\cite{leblebicioglu2026hybrid}, both planners have been modified and extended; these modifications are detailed in Sections~\ref{sec4.1} and~\ref{sec4.2}.

\subsection{Local Collision-Free Planning with enhanced RRT}\label{sec4.1}
In the first approach, RRT is used as the HLP together with the PF method. RRT is a probabilistic algorithm that incrementally builds a tree from the start toward the goal by randomly sampling points in the workspace~\cite{lavalle2001randomized}. For each sampled point, the nearest node in the existing tree is identified, and the tree is extended toward the sampled point if the path segment is collision-free. Unlike grid-based planners such as A*, RRT does not require an explicit model of the entire configuration space, which allows it to efficiently solve path planning problems in complex environments~\cite{zhang2018path}. If the start or goal node of the RRT lies within a rectangular or elliptical obstacle, it is displaced in the direction of $\mathbf{F}_{repulsive}$ using the RRT step size ($\eta$) until a collision-free position is obtained. The local target of the RRT is defined within a local horizon $L$ in the direction of $\mathbf{F}_{total}$ from the current robot position $\mathbf{q}$, as given in Eq.~\ref{eq:RRT}.

%This process is repeated until the goal is reached.

\begin{equation}
\label{eq:RRT}
\mathbf{q}_{\textit{target\_local}} =
\mathbf{q} + L
\frac{\mathbf{F}_{total}}
{\|\mathbf{F}_{total}\|}
\end{equation}

\begin{figure}
\centering
\includegraphics[width=0.65\textwidth]{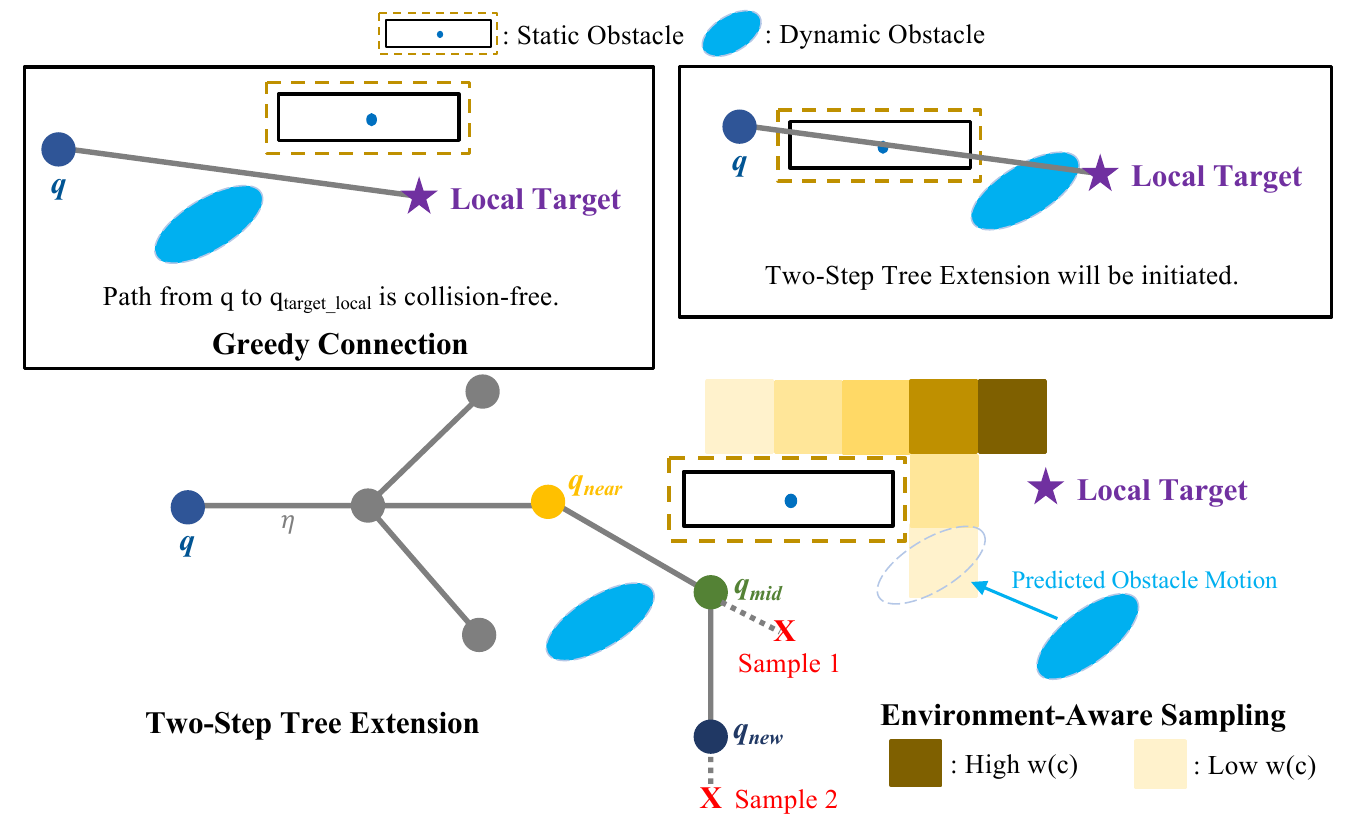}
\caption{Overview of the RRT enhancements. If the straight-line segment from the current robot position $\mathbf{q}$ to the local target $\mathbf{q}_{\textit{target\_local}}$ is collision-free, the planner returns it directly without building a tree. Otherwise, it builds the RRT tree using two-step extensions and environment-aware sampling.}
\label{fig:rrt_enhancements}
\end{figure}

In this work, the RRT planner is enhanced in three ways: a greedy direct connection, a two-step tree extension, and an environment-aware sampling distribution. Figure~\ref{fig:rrt_enhancements} provides a schematic overview of the enhancements introduced to the RRT planner.

\textbf{Greedy Direct Connection:} Since the RRT tree is rebuilt from scratch at every time step and only the first node of the resulting path is followed, the robot deviates in a different direction at each step, even in free space. To prevent this, before any tree expansion, the straight segment from $\mathbf{q}$ to $\mathbf{q}_{\textit{target\_local}}$ is checked for collisions. If it is collision-free, the path $[\mathbf{q}, \mathbf{q}_{\textit{target\_local}}]$ is returned directly without forming a tree, inspired by the greedy connection heuristic of RRT-Connect~\cite{kuffner2000rrt}. 

%As a result, the robot follows a zigzag trajectory, which is both slower and longer.
%with all rectangular and elliptical obstacles.
%Thus, a smoother path is obtained in free space. 

\textbf{Two-Step Tree Extension:} In each iteration, the tree is extended by two consecutive steps. First, the nearest node $\mathbf{q}_{near}$ is extended by a step of size $\eta$ toward the first sample, which gives an intermediate node $\mathbf{q}_{mid}$. Then, $\mathbf{q}_{mid}$ is extended by another step of size $\eta$ toward a second, independently drawn sample, which gives $\mathbf{q}_{new}$. Both nodes are added to the tree only if both segments are collision-free. Since the two steps generally point in different directions, a branch can bend within a single iteration, which helps the tree turn around obstacle corners and pass through narrow gaps. 
%instead of one.

\textbf{Environment-Aware Sampling:} Instead of uniform sampling, samples are drawn from a local grid of $10 \times 10$~m with a cell size of $0.5$~m. Each cell $c$ is assigned a weight $w(c)$, which is defined in Eq.~\ref{eq:rrt_weight} as the product of four terms. The term $w_{clear}$ favors cells far from the closest pedestrian, $w_{crowd}$ favors cells away from crowded regions, $w_{dir}$ penalizes cells that pedestrians are moving into, predicted from their current velocities over a short time horizon, and $w_{goal}$ favors cells close to the local target. Each term takes values in $[0,1]$, and cells inside obstacles are assigned zero weight.

%centered at the midpoint of $\mathbf{q}$ and $\mathbf{q}_{\textit{target\_local}}$.
%by accounting for all nearby pedestrians rather than only the closest one.

\begin{equation}
\label{eq:rrt_weight}
w(c) = w_{clear}(c) \times w_{crowd}(c) \times w_{dir}(c) \times w_{goal}(c)
\end{equation}

Since the weights are not probabilities by themselves, they are normalized to obtain the sampling probability of each cell, as given in Eq.~\ref{eq:rrt_prob}. By accounting for pedestrians' current positions and directions of motion, the weights guide the RRT toward more socially appropriate paths that avoid passing in front of dynamic obstacles.

\begin{equation}
\label{eq:rrt_prob}
P(c) = \frac{w(c)}{\sum_{c'} w(c')}
\end{equation}

\subsection{Local Collision-Free Planning with modified A*}\label{sec4.2}
In the second approach, A* is used as the HLP together with the PF method in the proposed hybrid planner. A* is a heuristic search algorithm designed to find the optimal path. The method is grid-based, where the environment is divided into small square cells (a grid), and the optimal path is determined by evaluating the total cost function $f(n)$ for neighboring nodes while progressing toward the goal, as given in Eq.~\ref{eq:Astar}. The A* algorithm maintains two values to calculate the total cost $f(n)$: $g(n)$ and $h(n)$, where $n$ represents the current node in the search. The term $g(n)$ denotes the cost from the start node to the current node, while $h(n)$ represents the heuristic estimate of the cost from the current node to the goal. During the search process, the node with the smallest total cost $f(n)$ is selected from the open set and expanded until the goal node is reached~\cite{huang2025self}.

\begin{equation}
\label{eq:Astar}
f(n) = g(n) + h(n)
\end{equation}

Similar to the RRT in Section~\ref{sec4.1}, the local target $\mathbf{q}_{\textit{target\_local}}$ is generated within the local horizon $L$ in the direction of $\mathbf{F}_{total}$ (Eq.~\ref{eq:RRT}), and A* plans a path from $\mathbf{q}$ to $\mathbf{q}_{\textit{target\_local}}$ at every time step. The search space is restricted to a local region centered at the current robot position, with search bounds $x \in [q_x-5, q_x+5]$ and $y \in [q_y-5, q_y+5]$. This region is represented on a grid with a resolution of 0.3~m, and $h(n)$ is defined as the Euclidean distance to the local target. If the start node or the local target lies inside an obstacle, it is pushed out of the obstacle along $\mathbf{F}_{repulsive}$, as in the RRT. Paths that intersect a rectangular or an elliptical obstacle are discarded. In this work, the A* planner is modified in two ways: an extended 24-neighborhood and a direction-dependent motion cost.

% If the start node or the local target lies inside an obstacle, it is displaced in the direction of $\mathbf{F}_{repulsive}$, as in the RRT. 

\begin{figure}
\centering
\includegraphics[width=0.5\textwidth]{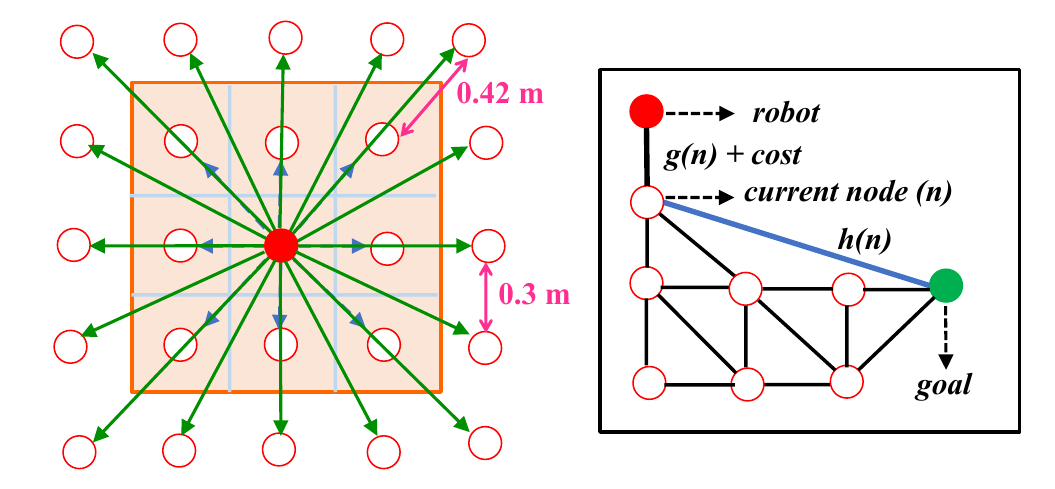}
\caption{Left: The 24-neighborhood used in A*: the 8 adjacent cells (blue) and the 16 cells two steps away (green), all reached directly from the current node, with a grid resolution of 0.3~m. Right: Computation of the total cost $f(n)=g(n)+h(n)$, where $g(n)$ includes the direction-dependent motion cost.}
\label{fig:Astar}
\end{figure}

\textbf{Extended 24-Neighborhood:} In~\cite{leblebicioglu2026hybrid}, each node had 8 neighbors, i.e., the adjacent cells in the horizontal, vertical, and diagonal directions. Here, the neighborhood is extended to the 24 cells within two cells of the current node, as shown in the left side of Fig.~\ref{fig:Astar}. Each of these neighbors is reached directly, with a single straight move whose cost is its Euclidean length, i.e., $0.3$~m for $(1,0)$, $0.42$~m for $(1,1)$, $0.6$~m for $(2,0)$, $0.67$~m for $(1,2)$, and $0.85$~m for $(2,2)$. Since a move can now skip over a cell, both endpoints of a move can be collision-free while the segment between them is not; therefore, every move is checked for collisions along the entire segment.

\textbf{Direction-Dependent Motion Cost:} To favor moves that continue in the current direction of the path and discourage moving backward, a penalty is added to the cost of each move according to the angle $\theta$ between the move and the direction in which the current node was entered. The cost of a move from $n$ to $n'$ is given in Eq.~\ref{eq:Astar_cost}, where $\Delta = 0.3$~m is the grid resolution, and $c_s = 1$ and $c_b = 100$ are the sideways and backward penalties, respectively. Hence, moving forward ($\theta = 0^\circ$) adds no penalty, so that the cost reduces to the length of the move; turning sideways ($\theta = 90^\circ$) is penalized by $c_s$; and moving backward ($\theta = 180^\circ$) is penalized by $c_b$. The cost of each move is then accumulated in $g$ as $g(n') = g(n) + c(n,n')$, while $h(n)$ is unchanged.

%To prevent the planned path from moving backward, i.e., in the direction opposite to its previous move

\begin{equation}
\label{eq:Astar_cost}
c(n,n') = \|n' - n\| + \Delta
\begin{cases}
c_s\,(1-\cos\theta), & \cos\theta \geq 0 \\[4pt]
c_s + (c_b - c_s)(-\cos\theta), & \cos\theta < 0
\end{cases}
\end{equation}

\subsection{Hybrid Path Construction for Multi-Goal Navigation}\label{4.3}
In dynamic environments, global and local planning are typically handled separately. In the proposed hybrid framework, the global planning problem is formulated as a TSP and solved using PSO. At the local level, robot motion is guided by the PF method together with RRT or A* to avoid collisions with both static and dynamic obstacles. Algorithm~\ref{alg:hybrid} presents the pseudocode for the two proposed approaches: (i) TSP-PF-RRT and (ii) TSP-PF-A*.
% In scenarios without stations, such as the Van-Ramp scenario (Sec.~\ref{sec8}), the TSP step reduces to a direct corner-routing path from the start to the goal, which is recomputed at every time step.

Since the local target of both HLPs is placed in the direction of $\mathbf{F}_{total}$ (Eq.~\ref{eq:RRT}), the direction of $\mathbf{F}_{rep}^{ellipse}$ determines how the wheelchair responds to a pedestrian. With a purely radial repulsive force, a pedestrian standing on the global path pushes the wheelchair directly backward, against its direction of travel, so that the wheelchair backs away and waits for the path to clear. To enable the wheelchair to maneuver around the pedestrian instead, a tangential component is added to the repulsive force of each dynamic obstacle, and the radial unit vector in Eq.~\ref{eq:Fellipse} is replaced by $\mathbf{d}_i$ as given in Eq.~\ref{eq:vortex}, where $\hat{\mathbf{r}}_i = (\mathbf{q}-\mathbf{o}_i)/\|\mathbf{q}-\mathbf{o}_i\|$ is the unit radial vector from the $i$-th dynamic obstacle to the wheelchair, and $\hat{\mathbf{p}}_i$ is $\hat{\mathbf{r}}_i$ rotated by $90^\circ$.

\begin{equation}
\label{eq:vortex}
\mathbf{d}_i = \frac{(1-k)\,\hat{\mathbf{r}}_i + k\,\hat{\mathbf{p}}_i}
{\|(1-k)\,\hat{\mathbf{r}}_i + k\,\hat{\mathbf{p}}_i\|}, \qquad
\mathbf{F}_{ell}^{(i)} =
\begin{cases}
k_{low}\, \mathbf{d}_i, & d_{safe} < d_E < d_0^{ell} \\[4pt]
k_{high}\, \mathbf{d}_i, & d_E \le d_{safe} \\[4pt]
0, & d_E \ge d_0^{ell}
\end{cases}
\end{equation}

Of the two possible perpendicular directions, the one satisfying $\hat{\mathbf{p}}_i \cdot \hat{\mathbf{t}} \geq 0$ is selected, where $\hat{\mathbf{t}}$ is the unit tangent of the global path at the point closest to the wheelchair, i.e., the same unit vector used to compute $\mathbf{F}_{tangent}$. The weight $k \in [0,1]$ balances the radial and tangential components; $k=0$ recovers Eq.~\ref{eq:Fellipse}, and $k=0.8$ is used in this work. As a result, the repulsive force steers the wheelchair around the pedestrian while keeping it moving forward along the global path. In addition, for both PF-RRT and PF-A*, if the HLP cannot find a path to the local target, the wheelchair waits at its current position until a collision-free path becomes available, instead of being driven by the PF alone.

\begin{algorithm}[t]
\caption{Hybrid TSP--PF--RRT or TSP--PF--A* Planner}
\label{alg:hybrid}
\begin{algorithmic}[1]

\State Solve TSP using PSO to obtain the global waypoint order
\State Construct the global geometric path $\mathcal{G}$ and polyline
\State $\mathbf{q}_{goal} \gets$ next waypoint on $\mathcal{G}$

\While{$\mathbf{q} \neq \mathbf{q}_{goal\_final}$}

\State Every 10~s, recompute $\mathcal{G}$ using PSO from $\mathbf{q}$ to $\mathbf{q}_{goal\_final}$ through the remaining waypoints
\State Apply stochastic maneuver to obstacle velocities
\State Update obstacle positions
\State $\hat{\mathbf{t}} \gets$ unit tangent of $\mathcal{G}$ at the point closest to $\mathbf{q}$
\State Compute $\mathbf{F}_{repulsive}$ with the tangential component (Eq.~\ref{eq:vortex})
\State Compute $\mathbf{F}_{total}$ using $\mathbf{F}_{attractive}$, $\mathbf{F}_{repulsive}$, $\mathbf{F}_{tangent}$ and $\mathbf{F}_{pull}$
\State $\mathbf{d} \gets \mathbf{F}_{total}/\|\mathbf{F}_{total}\|$
\State $\mathbf{q}_{\textit{target\_local}} \gets \mathbf{q} + L\mathbf{d}$
\State Plan path from $\mathbf{q}$ to $\mathbf{q}_{\textit{target\_local}}$ using RRT or A*
\If{a path is found}
    \State $\mathbf{q}_{next} \gets$ next node of the RRT or A* path
    \State $\mathbf{q} \gets \mathbf{q} + v_{robot}\, dt\, \frac{\mathbf{q}_{next}-\mathbf{q}}{\|\mathbf{q}_{next}-\mathbf{q}\|}$
\Else
    \State Wait at $\mathbf{q}$
\EndIf
\State Check collisions with obstacles
\If{current waypoint reached}
    \State $\mathbf{q}_{goal} \gets$ next waypoint
\EndIf

\EndWhile

\end{algorithmic}
\end{algorithm}

\section{Simulation-Based Evaluation of Point and Elliptical Wheelchair Models}\label{sec5}
The simulation environment is defined over the range $[-50,50]$~m in both the $x$ and $y$ directions. The initial position of the wheelchair is $\mathbf{q}_0 = [-40,-40]$, and the goal position is $\mathbf{q}_{goal} = [20,10]$. The simulation uses a fixed, user-defined time step of $\Delta t = 0.01$~s, independent of the planners' computation time, and the wheelchair moves with a constant speed of $v_r = 30$~m/s. For a fair comparison, both local planners use the same local horizon of $L = 1.0$~m, and the RRT step size is set equal to the A* grid resolution, i.e., $0.3$~m.

%The simulation time step is $\Delta t = 0.01$~s, and the wheelchair moves with a constant speed of $v_r = 30$~m/s.

\begin{figure}
\centering
\includegraphics[width=0.5\textwidth]{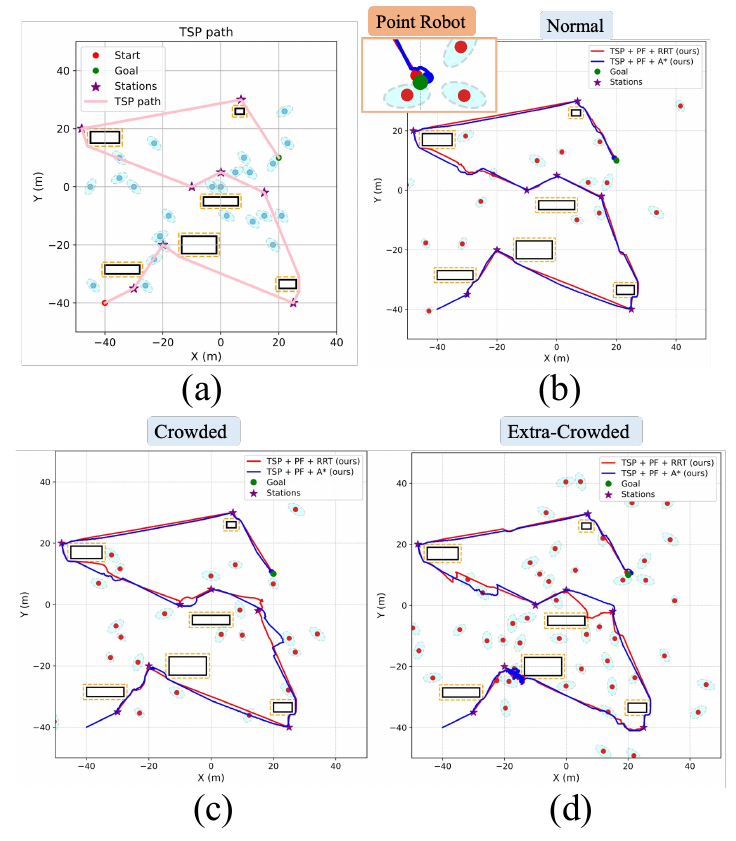}
\caption{(a) Global path obtained by the TSP. (b)--(d) Trajectories of TSP-PF-RRT (red) and TSP-PF-A* (blue) for the point model in the normal, crowded, and extra-crowded scenarios, respectively. The avoidance maneuvers increase as the scenario complexity increases.}
\label{fig:robot_point}
\end{figure}

The proposed methods described in Section~\ref{sec4} are compared through Monte Carlo (MC) simulations, where TSP is used for global planning and (i) PF-RRT and (ii) PF-A* are used for local obstacle avoidance. MC simulations are employed because the motions of the dynamic obstacles are random. The wheelchair is modeled both as a point and as an ellipse, and each model is evaluated in two scenarios, normal and crowded, with ten simulations for each combination. The normal and crowded scenarios contain 16 and 24 dynamic obstacles, respectively, and there are six static rectangular obstacles in all scenarios. In all cases, both methods follow the same global path, obtained by solving the TSP with PSO for an eight-waypoint trajectory, which is 271.9~m long (see Fig.~\ref{fig:robot_point}a). A PF-only planner, i.e., PF without RRT or A*, is not included in the comparison, since it cannot effectively avoid collisions in crowded environments.

The average total distance traveled and the average number of collisions over the ten trials are reported in Tables~\ref{tab:comparison1} and~\ref{tab:comparison2} for the point and elliptical models, respectively. Note that collisions are counted at every time step, based on the noisy obstacle positions perceived by the wheelchair. Hence, a single encounter that lasts several time steps is counted as multiple collisions, and the reported number reflects the duration of contact rather than the number of distinct collision events. In all trials, collisions occur only with dynamic obstacles. 

\textbf{Point Model:} As shown in Table~\ref{tab:comparison1}, PF-RRT does not collide in any trial, in both the normal and crowded scenarios, while PF-A* has 6.1 and 18.4 collisions on average, respectively. The distances traveled by the two methods are similar in both scenarios. In the crowded scenario, both methods perform more maneuvers around the pedestrians (Fig.~\ref{fig:robot_point}c), so that the distance traveled increases by approximately 20\% for both methods, and the number of collisions of PF-A* triples. To further test the limits of both planners, an extra-crowded scenario with 56 dynamic obstacles is also evaluated for the point model only, since the gaps between the pedestrians in this scenario are often narrower than the elliptical wheelchair body. In a representative trial, shown in Fig.~\ref{fig:robot_point}d, PF-RRT again reaches the goal without any collision, traveling 405.2~m, while PF-A* travels 396.6~m with 53 collisions.

%In the normal scenario, both distances are close to the length of the global path.

\begin{figure}
\centering
\includegraphics[width=0.5\textwidth]{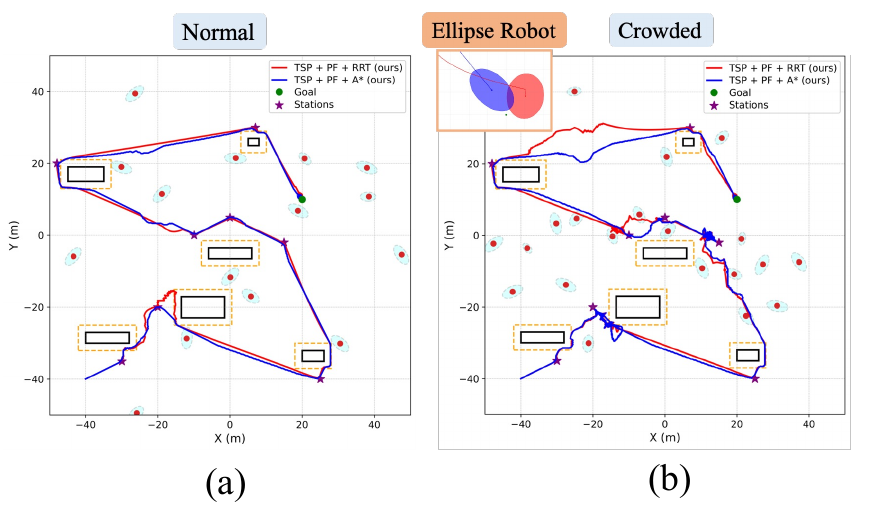}
\caption{Trajectories of TSP-PF-RRT (red) and TSP-PF-A* (blue) for the elliptical model in the normal (left) and crowded (right) scenarios.}
\label{fig:robot_ellipse}
\end{figure}

\textbf{Elliptical Model:} When the wheelchair is modeled as an ellipse, both methods collide with pedestrians even in the normal scenario, as shown in Table~\ref{tab:comparison2}. In this case, the two methods show similar performance: the distances traveled are almost identical, and PF-RRT has slightly fewer collisions than PF-A* (14.3 vs. 19.0 in the normal scenario). In the crowded scenario, PF-RRT travels approximately 8~m (2\%) more than PF-A* (400.0 vs. 392.0~m) but has 33\% fewer collisions (17.0 vs. 25.3), which can be seen as the cost of its more extensive avoidance maneuvers (Fig.~\ref{fig:robot_ellipse}).

%since the wheelchair body occupies a larger area than a single point. 
% Compared to the point model in the normal scenario, the distance traveled increases by approximately 11\%, since the elliptical body requires wider maneuvers around the pedestrians (Fig.~\ref{fig:normal_ellipse}), and the number of collisions of PF-A* increases by a factor of three. In the crowded scenario, the distance traveled increases further, by approximately 30\% compared to the normal scenario for both methods, and the number of collisions increases by 19\% for PF-RRT and 33\% for PF-A* (Fig.~\ref{fig:robot_ellipse}). PF-RRT again has fewer collisions than PF-A*.

In all scenarios, both methods maneuver around the pedestrians instead of backing away from them, owing to the tangential component of the repulsive force (Eq.~\ref{eq:vortex}). After each maneuver, the wheelchair returns to the global path, since $\mathbf{F}_{pull}$ attracts it back toward the path and $\mathbf{F}_{tangent}$ keeps it moving along the path direction.

\begin{table}[t]
\centering
\caption{Average distance traveled and number of collisions over 10 MC trials in the normal and crowded scenarios for the point model.}
\label{tab:comparison1}
\begin{tabular}{lcccc}
\hline
Method & \multicolumn{2}{c}{Normal} & \multicolumn{2}{c}{Crowded} \\
 & Distance (m) & Collisions & Distance (m) & Collisions \\
\hline
TSP--PF--RRT & 273.0 & 0 & 332.0 & 0 \\
TSP--PF--A*  & 277.6 & 6.1 & 330.7 & 18.4 \\
\hline
\end{tabular}
\end{table}

\begin{table}[t]
\centering
\caption{Average distance traveled and number of collisions over 10 MC trials in the normal and crowded scenarios for the elliptical model.}
\label{tab:comparison2}
\begin{tabular}{lcccc}
\hline
Method & \multicolumn{2}{c}{Normal} & \multicolumn{2}{c}{Crowded} \\
 & Distance (m) & Collisions & Distance (m) & Collisions \\
\hline
TSP--PF--RRT & 306.4 & 14.3 & 400.0 & 17.0 \\
TSP--PF--A*  & 306.3 & 19.0 & 392.0 & 25.3 \\
\hline
\end{tabular}
\end{table}

% \begin{figure}
% \centering
% \includegraphics[width=0.75\textwidth]{figures/point_robot.pdf}
% \caption{Trajectories of TSP-PF-RRT (red) and TSP-PF-A* (blue) for the point model in the normal (left) and crowded (right) scenarios.}
% \label{fig:robot_point}
% \end{figure}

% \begin{figure}
% \centering
% \includegraphics[width=0.75\textwidth]{figures/extra_crowded.pdf}
% \caption{Left: The global path obtained by the TSP. Right: trajectories of TSP-PF-RRT (red) and TSP-PF-A* (blue) for the point model in the extra-crowded scenario.}
% \label{fig:extra_crowded}
% \end{figure}

% \begin{figure}
% \centering
% \includegraphics[width=0.75\textwidth]{figures/normal_ellipse.pdf}
% \caption{Left: The global path obtained by the TSP. Right: Trajectories of TSP-PF-RRT (red) and TSP-PF-A* (blue) for the elliptical model in the normal scenario.}
% \label{fig:normal_ellipse}
% \end{figure}

% \begin{figure}
% \centering
% \includegraphics[width=0.75\textwidth]{figures/robot_ellipse.pdf}
% \caption{Left: The global path obtained by the TSP. Right: Trajectories of TSP-PF-RRT (red) and TSP-PF-A* (blue) for the elliptical model in the crowded scenario.}
% \label{fig:robot_ellipse}
% \end{figure}

In summary:
\begin{itemize}
    \item PF-RRT and PF-A* travel similar distances in all scenarios, for both the point and elliptical models.
    \item For both methods, the distance traveled increases as the environment becomes more crowded and when the wheelchair is modeled as an ellipse instead of a point.
    \item With the point model, PF-RRT avoids all collisions, even in the extra-crowded scenario, whereas the collisions of PF-A* increase with the number of dynamic obstacles.
    \item With the elliptical model, the collisions of PF-A* increase further, and PF-RRT also starts to collide, but still less than PF-A*.
\end{itemize}
These results indicate that, while both methods are comparable in path efficiency (i.e., distance traveled), PF-RRT is more effective in avoiding collisions with dynamic obstacles, particularly in crowded environments. This is likely due to its sampling-based exploration: by randomly extending the tree in multiple directions, RRT can find collision-free gaps between moving pedestrians that the grid-based search of A* may miss, consistent with the findings in~\cite{leblebicioglu2026hybrid}.

\section{Physical Wheelchair Platform}\label{sec6}
\begin{figure}[h]
\centering
\includegraphics[width=0.5\textwidth]{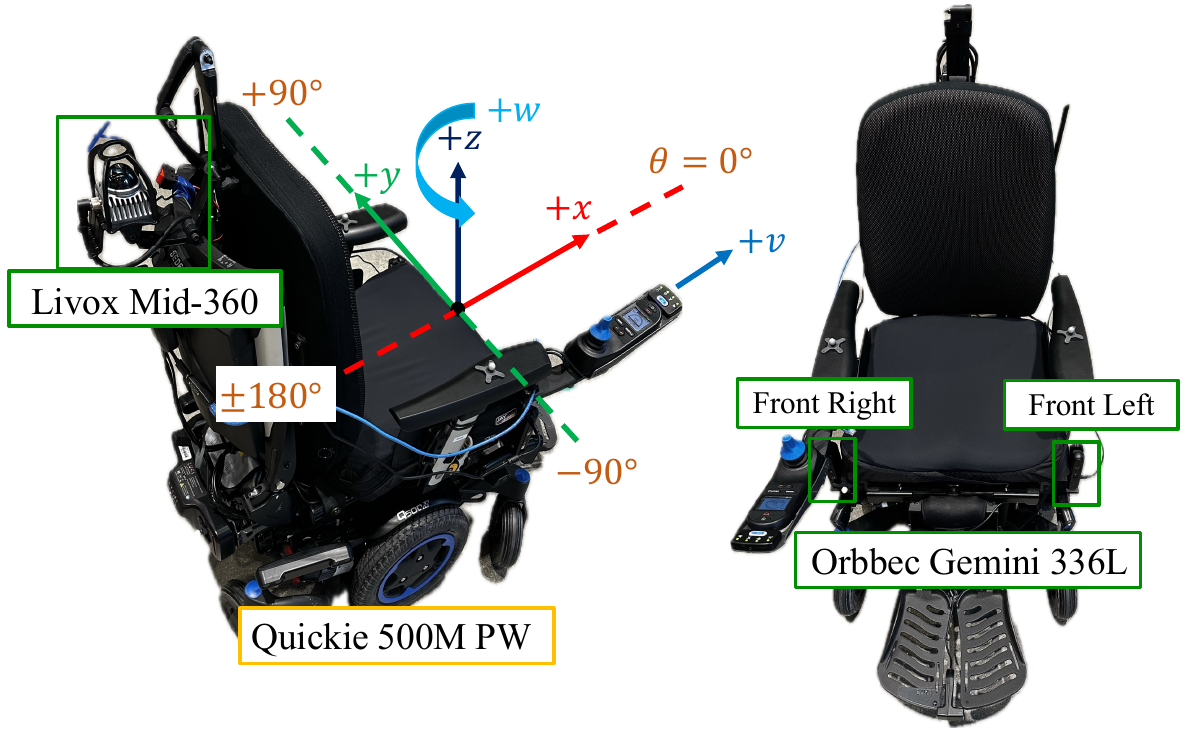}
\caption{Wheelchair body coordinate frame following the ROS REP-103 convention, showing the positive directions of linear velocity, angular velocity, and yaw, along with the onboard sensor locations.}\label{fig:physical_wheelchair}
\end{figure}

The physical platform used in this study is a Quickie 500M PW equipped with the LUCI mobility system and running the LUCI ROS 2 SDK~\cite{luci_sdk_docs}. LUCI provides a collision-avoidance and safety layer through sensor fusion. In this work, the wheelchair is operated in autonomous remote mode with the LUCI safety layer fully disabled, so that obstacle avoidance is achieved solely by the proposed algorithms. Coordinate frames follow the right-handed ROS REP-103 convention~\cite{rep103}, with the $x$-axis pointing forward, the $y$-axis to the left, and the $z$-axis upward in the body frame. The heading (yaw) angle, $\theta$, is measured relative to the global $+x$-axis and increases counter-clockwise (CCW). Left and right turns correspond to positive and negative angular velocity ($\omega$), respectively, while forward and reverse motion correspond to positive and negative linear velocity ($v$), respectively (see Fig.~\ref{fig:physical_wheelchair}). All experiments were conducted using speed profile 3, for which the maximum speed is 3.6 mph (approximately 1.6 m/s).
% combining Intel RealSense D430 depth cameras and radar sensors at the front and rear with ultrasonic sensors along the sides and at the corners. The depth cameras detect abrupt changes in ground elevation, including drop-offs, depressions, and raised surfaces; radar sensor detects moving and large obstacles; and ultrasonic sensors provide coverage in blind spots.
%with $0^\circ$ along $+x$, $90^\circ$ along $+y$, and $\pm180^\circ$ along $-x$.
%The chair provides five speed profiles.

The wheelchair accepts joystick commands with forward--back (FB) and left--right (LR) components bounded to $[-100,100]$. The physical limitations of the wheelchair influence the design of the local planner. In particular, the joystick input has a deadzone: commands below a minimum magnitude produce no motion. A reference point too close to the wheelchair can therefore generate a command within this deadzone. Consequently, unlike in simulation, the first node of the planned path is not used as the reference point. For the TSP-PF-A* planner, which uses a grid resolution of 0.15 m in the hardware experiments, the reference point is selected at the third node, approximately 0.45 m along the path. This selection produces commands above the deadzone threshold and eliminates the left--right oscillations observed with closer reference points. For the TSP-PF-RRT planner, the reference point is selected at a fixed look-ahead distance of 0.6 m, measured by arc length along the smoothed path, because variable node spacing makes index-based selection unreliable. For hardware experiments conducted outside the MoCap, two Orbbec Gemini 336L RGB-D cameras~\cite{orbbec_gemini336l} mounted at the front left and front right of the wheelchair are used for pedestrian tracking, while a Livox Mid-360 LiDAR~\cite{livox_mid360} mounted at the rear is used for localization. Within the MoCap coverage area, pedestrian tracking and wheelchair localization rely on marker-based MoCap data. 

%The mapping from controller outputs to joystick commands is detailed in Section~\ref{sec6.1}.
%These sensors are independent of the LUCI sensor-fusion kit and are used exclusively by the proposed pipeline, with the LUCI safety layer disabled throughout the experiments.

\subsection{Control Structure of the Physical Wheelchair}\label{sec6.1}

The proposed path-planning algorithms, TSP--PF--RRT and TSP--PF--A*, serve as guidance algorithms that generate a reference position for the wheelchair. Three proportional-integral-derivative (PID) controllers are implemented: a position controller ($\mathrm{PID}_{\mathrm{position}}$), consisting of separate controllers for the $x$- and $y$-axes ($\mathrm{PID}_{\mathrm{x}}$ and $\mathrm{PID}_{\mathrm{y}}$, respectively), and a heading controller ($\mathrm{PID}_{\theta}$). The path-planning and control algorithms are executed within the same control loop at 20 Hz, corresponding to a sampling interval of $\Delta t = 0.05~\mathrm{s}$. The discrete time instants are therefore defined as $t_k = k\Delta t$, where $k = 0, 1, 2, \ldots$ and the wheelchair position at $t_k$ is denoted by $x_k = x(t_k)$ and $y_k = y(t_k)$. At each time step $k$, the current position of the wheelchair, obtained from either the MoCap or the LiDAR-based navigation, is represented by the position vector $\mathbf{p}_{\mathrm{meas},k}=[x_k, y_k]^{T}$. The reference position is $\mathbf{p}_{\mathrm{ref},k}=[x_k^{\mathrm{ref}}, y_k^{\mathrm{ref}}]^{T}$. The position error vector is $\mathbf{e}_k = [e_{x,k},\, e_{y,k}]^T = \mathbf{p}_{\mathrm{ref},k} - \mathbf{p}_{\mathrm{meas},k}$, where $e_{x,k}$ and $e_{y,k}$ are the inputs to the $\mathrm{PID}_{\mathrm{x}}$ and $\mathrm{PID}_{\mathrm{y}}$ controllers, respectively. The outputs of $\mathrm{PID}_{\mathrm{x}}$ and $\mathrm{PID}_{\mathrm{y}}$ controllers are reference velocities along $x$- and $y$-axes, denoted by $v_{x,k}^{\mathrm{ref}}$ and $v_{y,k}^{\mathrm{ref}}$, respectively. The corresponding reference linear speed is calculated by taking the Euclidean norm of these velocity components, i.e., $v_k^{\mathrm{ref}} = \sqrt{(v_{x,k}^{\mathrm{ref}})^2 + (v_{y,k}^{\mathrm{ref}})^2}$. The reference heading angle is determined from the direction of the reference velocity vector as $\theta_k^{\mathrm{ref}}=\operatorname{atan2}\left(v_{y,k}^{\mathrm{ref}},v_{x,k}^{\mathrm{ref}}\right)$. The heading error, calculated as $e_{\theta,k} = \operatorname{wrap}_{[-\pi,\pi]}\left(\theta_k^{\mathrm{ref}} - \theta_{\mathrm{meas},k}\right)$, where $\theta_{\mathrm{meas},k}$ is the measured wheelchair heading, is provided as the input to $\mathrm{PID}_{\theta}$, which generates the angular velocity command $\omega_{\mathrm{PID},k}$.

% \begin{equation}
% \mathbf{e}_k =
% \begin{bmatrix}
% e_{x,k} \\
% e_{y,k}
% \end{bmatrix}
% =
% \begin{bmatrix}
% x_{k}^{\mathrm{ref}} \\
% y_{k}^{\mathrm{ref}}
% \end{bmatrix}
% -
% \begin{bmatrix}
% x_{k} \\
% y_{k}
% \end{bmatrix},
% \label{eq:error}
% \end{equation}

%The measured wheelchair heading $\theta_{\mathrm{meas},k}$, is obtained from the MoCap or LiDAR-based navigation.

The wheelchair speed at each time step $k$ is estimated from the consecutive position measurements provided by either the MoCap or LiDAR-based navigation. First, the raw speed ($v_{\mathrm{raw},k}$) is computed using a backward finite-difference approximation, as given in Eq.~\ref{eq:v_raw}. Since numerical differentiation may introduce high-frequency noise, the resulting speed estimate $v_{\mathrm{raw},k}$ is passed through a low-pass filter (LPF), as defined in Eq.~\ref{eq:v_filter}, to obtain the filtered measured speed $v_{\mathrm{meas},k}$. The resulting $v_{\mathrm{meas},k}$ is then used to scale the angular velocity $\omega_{\mathrm{PID},k}$ generated by the heading controller according to Eq.~\ref{eq:joystick_scaling}. As the wheelchair speed increases, the commanded angular velocity decreases, resulting in a smoother turning behavior. The scaled angular velocity $\omega_{\mathrm{cmd},k}$ is subsequently limited to $\pm\omega_{\max}$, where $\omega_{\max}=0.8~\mathrm{rad/s}$. 

%This limits aggressive maneuvers and abrupt yaw changes at high speeds. It will be converted to the corresponding LR LUCI2 joystick command. The measured speed is initialized as $v_{meas,0} = 0$.

\begin{equation}
\begin{aligned}
\Delta x_k &= x_k-x_{k-1},
\qquad
\Delta y_k = y_k-y_{k-1}, \\[3pt]
v_{\mathrm{raw},k}
&=
\frac{\sqrt{(\Delta x_k)^2+(\Delta y_k)^2}}{\Delta t},
\qquad k \geq 1
\end{aligned}
\label{eq:v_raw}
\end{equation}

\begin{equation}
v_{\mathrm{meas},k}
=
0.7\,v_{\mathrm{meas},k-1}
+
0.3\,v_{\mathrm{raw},k},
\qquad k \geq 1
\label{eq:v_filter}
\end{equation}

\begin{equation}
\omega_{\mathrm{cmd},k}
=
\frac{\omega_{\mathrm{PID},k}}
{1+k_v v_{\mathrm{meas},k}},
\qquad k_v = 1.5
\label{eq:joystick_scaling}
\end{equation}

The linear speed $v_k^{\mathrm{ref}}$ is further adjusted according to $e_{\theta,k}$ to prevent the wheelchair from moving at a high linear speed when its orientation is substantially misaligned from $\theta_k^{\mathrm{ref}}$. The commanded linear speed $v_{\mathrm{cmd},k}$ is calculated using Eq.~\ref{eq:linear_velocity_scaling}, where $v_{\max} = 3.0~\mathrm{m/s}$ denotes the maximum allowable linear speed. 

%The commanded linear speed decreases as the heading error increases. The resulting $v_{\mathrm{cmd},k}$ is then converted to the corresponding FB LUCI2 joystick command.

\begin{equation}
v_{\mathrm{cmd},k}
=
\min
\left(
v_{\max},
\;
v_k^{\mathrm{ref}} \max\left(0,\cos(e_{\theta,k})\right)
\right),
\label{eq:linear_velocity_scaling}
\end{equation}

\begin{figure}[h]
\centering
\includegraphics[width=0.8\textwidth]{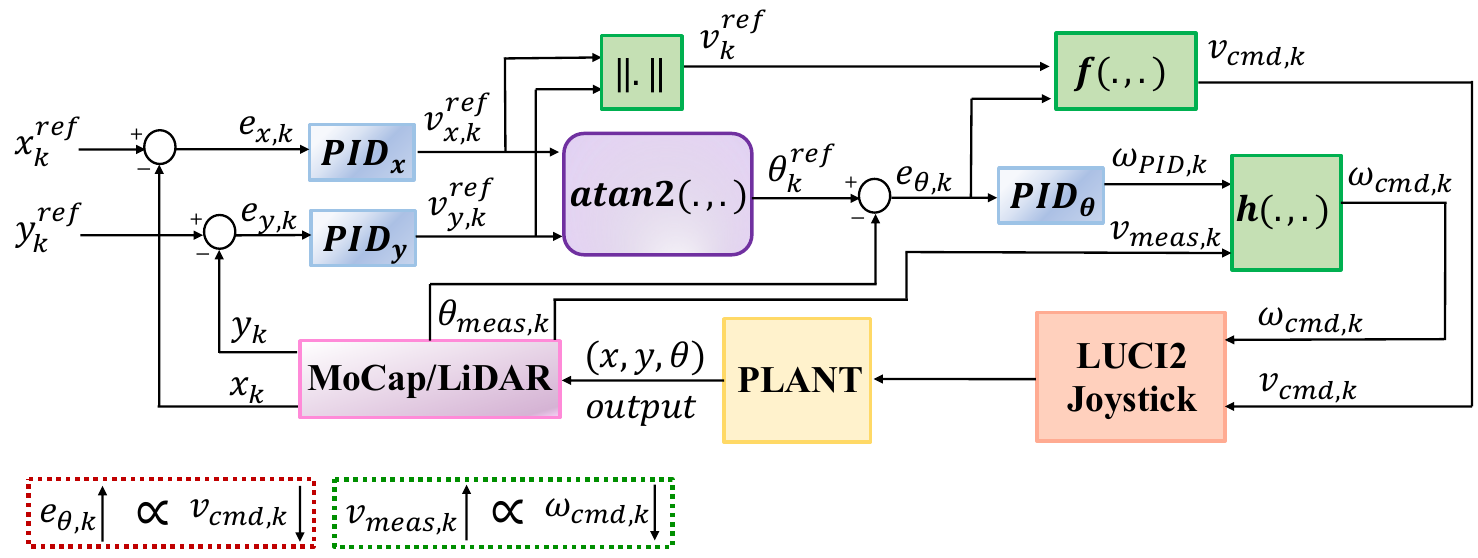}
\caption{Block diagram of the control architecture: the reference $\mathbf{p}_{\mathrm{ref},k}$ from the guidance algorithms is the input to the PID controllers, whose output is mapped to LUCI joystick commands. The blocks $f(\cdot,\cdot)$ and $h(\cdot,\cdot)$ correspond to Eqs.~\ref{eq:linear_velocity_scaling} and~\ref{eq:joystick_scaling}, respectively.}\label{fig:control}
\end{figure}

The LUCI2 joystick commands were obtained by experimentally fitting a curve for the measured wheelchair velocities to the corresponding joystick commands. The resulting FB and LR joystick commands are rounded to integer values, saturated to the allowable range of $[-100,\,100]$, and adjusted according to the corresponding joystick dead zones. The block diagram of the control architecture is given in Fig.~\ref{fig:control}. In the hardware experiments, reverse motion is disabled. When the reference point lies behind the wheelchair, the commanded linear velocity is zero (Eq.~\ref{eq:linear_velocity_scaling}, where $v_{\mathrm{cmd},k}=0$ for heading error magnitudes of $90^\circ$ or greater). The wheelchair therefore rotates in place to reduce the heading error before moving forward. This restriction is applied in both experimental settings to maintain consistency, since pedestrian tracking outside the MoCap area relies solely on the front left and front right cameras.

% The LUCI2 joystick commands corresponding to the commanded linear and angular velocities are determined using experimentally obtained velocity-to-joystick mappings for speed profile 3. These curves were obtained by fitting the measured wheelchair velocities to the corresponding joystick commands. The resulting mappings from $v_{\mathrm{cmd},k}$ and $\omega_{cmd,k}$ to the FB and LR joystick commands are given in Eqs.~\ref{eq:luci_fb} and~\ref{eq:luci_lr}, respectively. The resulting FB and LR joystick commands are rounded to integer values, saturated to the allowable range of $[-100,\,100]$, and adjusted according to the corresponding joystick dead zones.

% \begin{equation}
% \mathrm{FB}_k
% =
% 2.9095\,v_{\mathrm{cmd},k}^{3}
% -
% 3.4397\,v_{\mathrm{cmd},k}^{2}
% +
% 61.9742\,v_{\mathrm{cmd},k},
% \label{eq:luci_fb}
% \end{equation}

% \begin{equation}
% \mathrm{LR}_k
% =
% -\operatorname{sgn}\left(\omega_{\mathrm{cmd},k}\right)
% \frac{
% \left|\omega_{\mathrm{cmd},k}\right|-0.026822
% }{
% 0.003334
% },
% \label{eq:luci_lr}
% \end{equation}

\section{Hardware Experiments}\label{sec7}

\begin{figure}[h]
\centering
\includegraphics[width=0.75\textwidth]{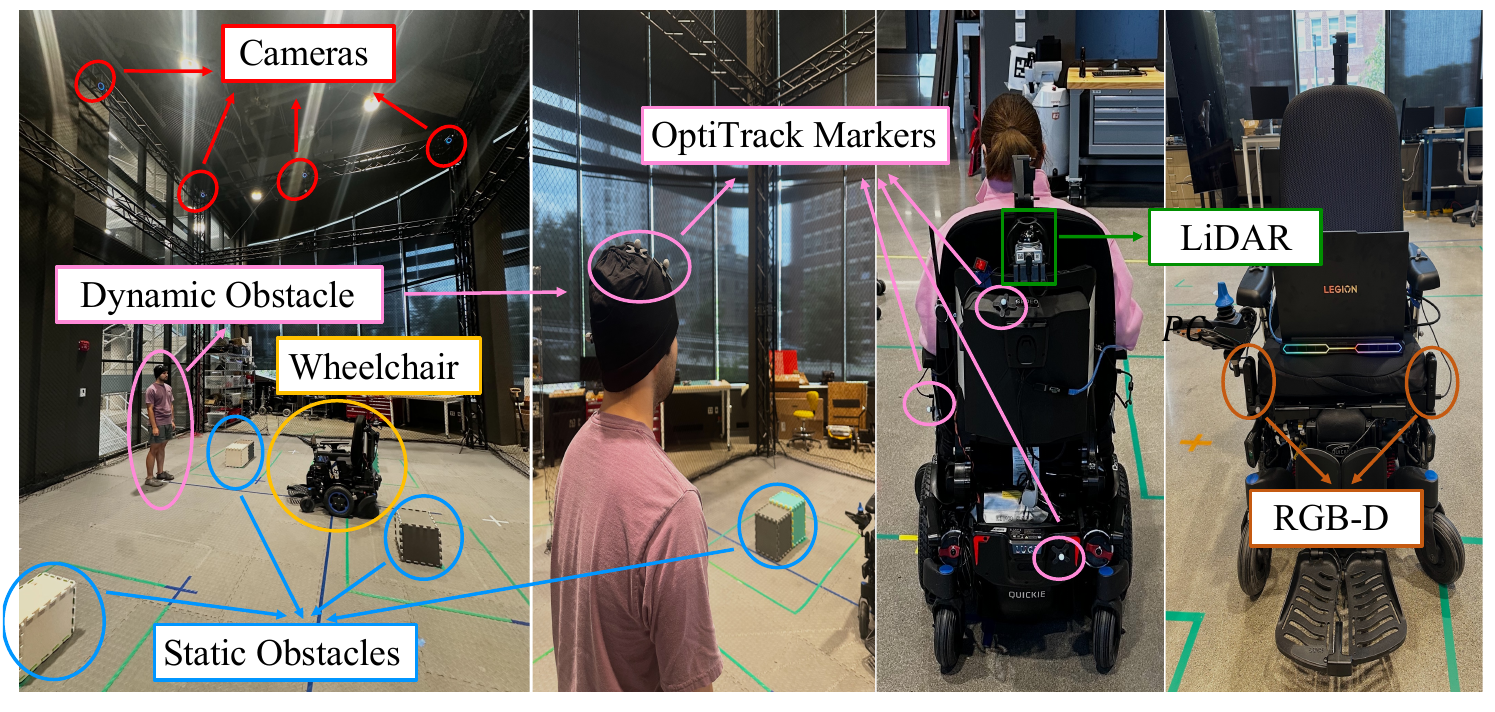}
\caption{Hardware experimental setup. First panel: the OptiTrack MoCap area. Second panel: Reflective markers attached to the participant's beanie and the wheelchair. Third and fourth panels: Sensor placement on the wheelchair, showing the rear-mounted LiDAR and the two front-mounted RGB-D cameras.}\label{fig:general} %with the wheelchair, a dynamic obstacle, and static obstacles
\end{figure}

Two sets of complementary hardware experiments were conducted to evaluate the proposed path planning framework. The first set of experiments was conducted in a $5.4~\mathrm{m} \times 5.3~\mathrm{m}$ cage equipped with an OptiTrack MoCap system. Reflective markers were mounted on the wheelchair as well as on the beanies that human participants wore. Measurements obtained from the MoCap provide ground-truth position and orientation information for both the wheelchair and the humans in the cage. The second set of experiments was conducted in a $7.8~\mathrm{m} \times 6.7~\mathrm{m}$ open area outside the MoCap, using onboard sensors. Static obstacles of different sizes were constructed by stacking foam cubes, each with a side length of 0.61~m (24~in). The extended rectangles and waypoints were marked on the floor using vinyl tape. The experimental setup is shown in Fig.~\ref{fig:general}.

%There are eight overhead cameras covering the cage. Rigid-body assets were created in Motive software to obtain the three-dimensional coordinates $(x,y,z)$ and orientations (roll, pitch, yaw) of the wheelchair and the dynamic obstacles (humans) inside the cage.
%to track their positions and orientations throughout the experiments.
%In this configuration, LiDAR-based navigation was used to estimate the wheelchair pose, while RGB-D cameras were used to detect and track humans and estimate their relative positions with respect to the wheelchair.
%into configurations of varying height and width.

\begin{figure}[h]
\centering
\includegraphics[width=0.75\textwidth]{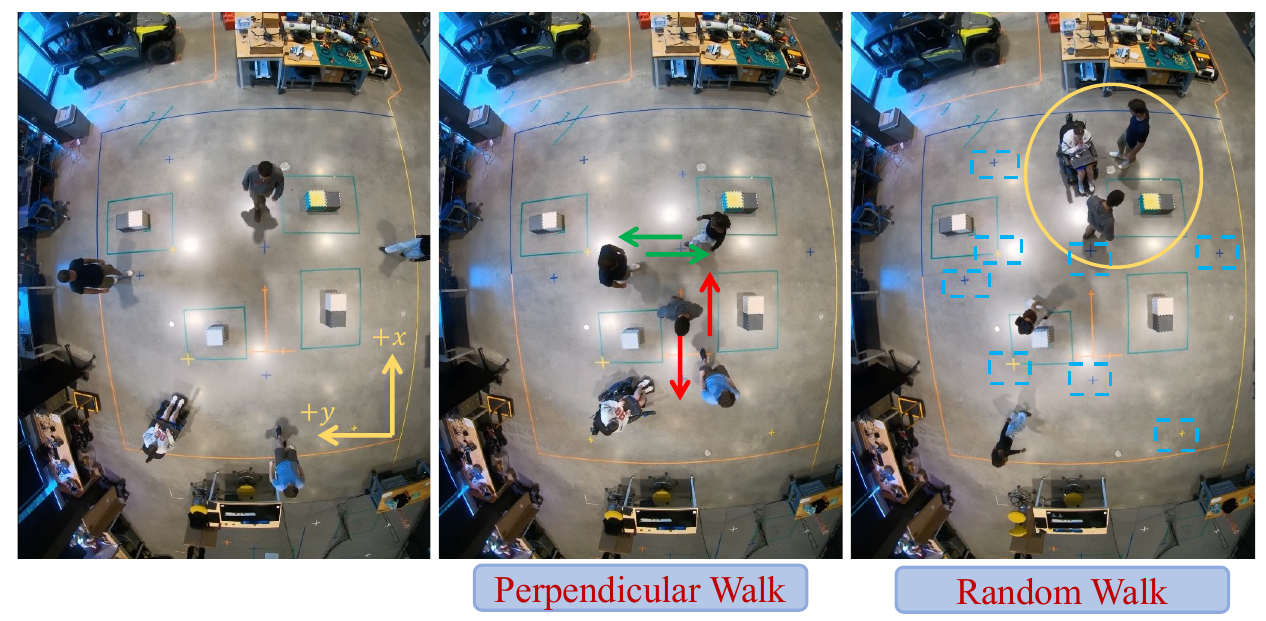}
% \caption{Overhead view of the experimental environment outside MoCap. Left: Test area with static obstacles (boxes inside green extended rectangles), the wheelchair, and pedestrians; the yellow arrows indicate the $+x$ and $+y$ axes. Middle: Perpendicular Walk Scenario: two pedestrians walk back and forth along the $x$-axis (red arrows) and two along the $y$-axis (green arrows). Right: Random Walk Scenario: two pedestrians move randomly while the other two attempt to block the wheelchair near the waypoints and rectangle corners (blue dashed boxes); the yellow circle highlights a blocking event.}
\caption{Overhead view of the experimental environment outside MoCap. Left: Test area with static obstacles (boxes inside green extended rectangles), the wheelchair, and pedestrians; the yellow arrows indicate the $+x$ and $+y$ axes. Middle: Perpendicular Walk scenario (red and green arrows: pedestrian directions). Right: Random Walk scenario (yellow circle: a blocking event).}
\label{fig:exps}
\end{figure}

Two sets of experiments are conducted for each algorithm in both experimental settings. The total distance traveled by the wheelchair is compared across algorithms. Each set involves four pedestrians: (A) Perpendicular Walk, in which two pedestrians walk back and forth along the $x$-axis while the other two walk back and forth along the $y$-axis (middle panel of Fig.~\ref{fig:exps}); and (B) Random Walk, in which two pedestrians move randomly while the other two actively attempt to block the wheelchair's path near the waypoints and rectangle corners (blue dashed boxes in the right panel of Fig.~\ref{fig:exps}). No collisions are reported with pedestrians in hardware experiments. 
%the MoCap-based setting and the LiDAR- and RGB-D-based setting outside the MoCap coverage area.

\subsection{MoCap-Based Experiments}\label{sec7.1}

\begin{figure}[h]
\centering
\includegraphics[width=0.75\textwidth]{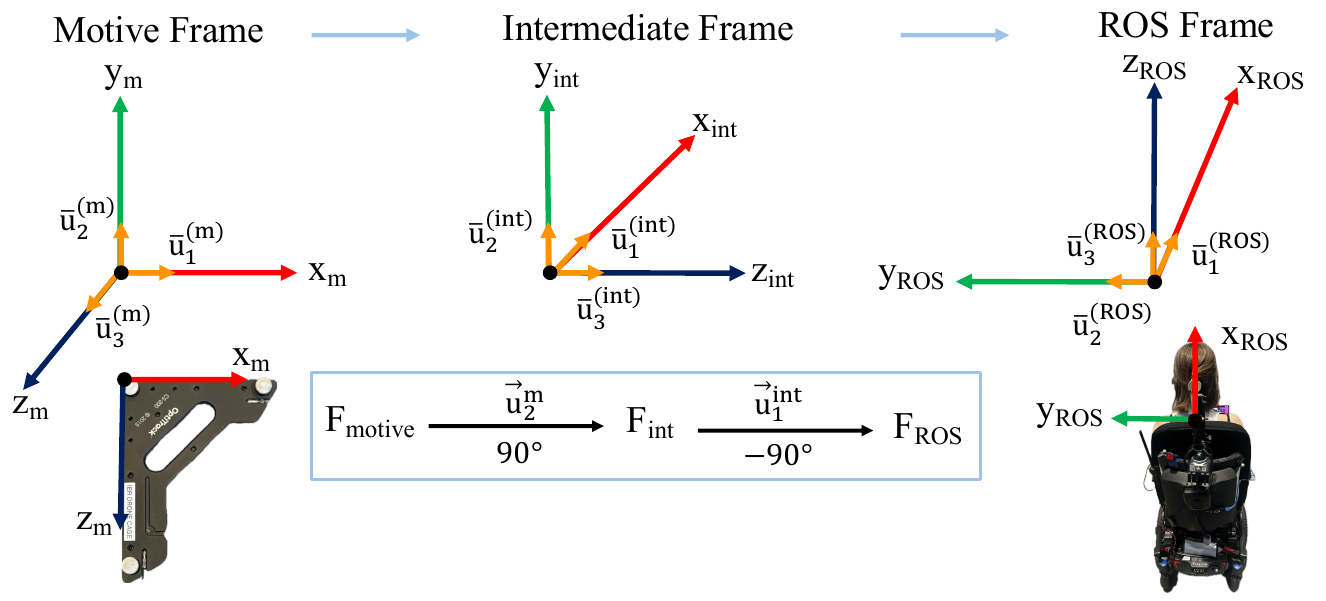}
\caption{Coordinate frame transformation between the Motive and ROS reference frames. The unit vectors of each axis are labeled as $\bar{u}_i^{\textit{(m, int, ROS)}}$,
where $i \in \{1,2,3\}$ denotes the axis ($x$, $y$, $z$). The short edge of the calibration triangle is the $x$-axis.}\label{fig:ref_frames}
\end{figure}

The MoCap system provides wheelchair pose measurements at a sampling frequency of 240 Hz. These measurements are continuously received by the ROS 2 node, while the most recent pose measurement is used by the path-planning and control loop, which operates at a frequency of 20 Hz. The rotation between the Motive reference frame $\mathbf{F}_{\text{motive}}$
(with axes $x_m$, $y_m$, $z_m$) and the ROS reference frame
$\mathbf{F}_{\text{ROS}}$ (with axes $x_{\text{ROS}}$, $y_{\text{ROS}}$,
$z_{\text{ROS}}$) is defined as
\begin{equation}
    \hat{C}^{(\text{motive},\,\text{ROS})} = \hat{R}_2(90^\circ)\,\hat{R}_1(-90^\circ),
\end{equation}
where $\hat{R}_i$ ($i = 1,2,3$) is the basic rotation matrix used to define
the basic rotation between two consecutive reference
frames. The first rotation is around the $y$-axis of the Motive reference frame
$\mathbf{F}_{\text{motive}}$ by $90^\circ$. The second
rotation is around the $x$-axis of the rotated reference frame, i.e., the
intermediate reference frame $\mathbf{F}_{\text{int}}$ (with axes
$x_{\text{int}}$, $y_{\text{int}}$, $z_{\text{int}}$), by $-90^\circ$. Both rotations follow the right-hand rule. Hence, the corresponding transformation matrix between the
$\mathbf{F}_{\text{motive}}$ and the $\mathbf{F}_{\text{ROS}}$ is given in Eq.~\ref{eq1} (see Fig.~\ref{fig:ref_frames}). This matrix maps vector components expressed in the ROS frame to those expressed in the Motive frame, yielding \(x_m=-y_{\mathrm{ROS}}\), \(y_m=z_{\mathrm{ROS}}\), and \(z_m=-x_{\mathrm{ROS}}\). The inverse mapping, obtained using the transpose of this matrix, yields \(x_{\mathrm{ROS}}=-z_m\), \(y_{\mathrm{ROS}}=-x_m\), and \(z_{\mathrm{ROS}}=y_m\), consistent with the coordinate mapping implemented in the ROS 2 driver node.

\begin{equation}
    \hat{C}^{(\text{motive},\,\text{ROS})} =
    \begin{bmatrix}
        0 & -1 & 0 \\
        0 & 0 & 1 \\
        -1 & 0 & 0
    \end{bmatrix}
    \label{eq1}
\end{equation}

% \begin{figure}[h]
% \centering
% \includegraphics[width=0.5\textwidth]{figures/global_path_mocap1.pdf}
% \caption{MoCap-based hardware experiment without dynamic obstacles.
% Left: Global path obtained by the TSP planner with a total length of 15~m.
% Right: The wheelchair trajectory (navy) recorded by MoCap while tracking
% the global path (pink) using only the PF method, with a total distance traveled
% of 14.4~m. Black rectangles are static rectangular obstacles and dashed orange
% rectangles are extended rectangles (buffer is 0.45~m).}\label{fig:globalpath_mocap}
% \end{figure}

\begin{figure}[h]
\centering
\includegraphics[width=1.0\textwidth]{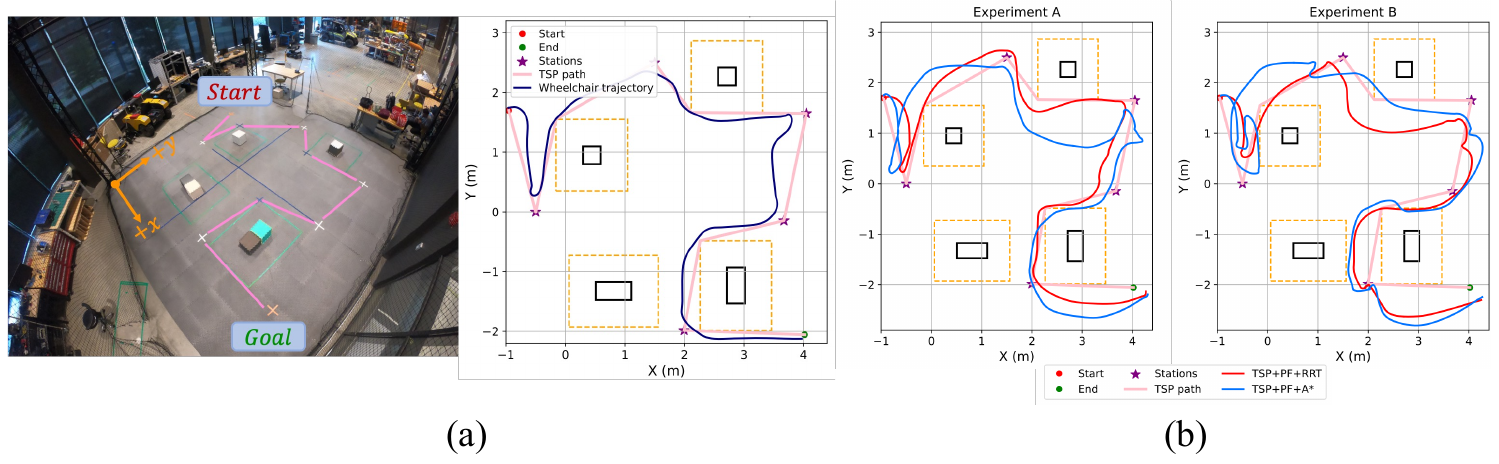}
\caption{MoCap-based hardware experiments. (a) Without dynamic obstacles. Left: Global path obtained by the TSP planner (15~m). Right: Wheelchair trajectory (navy) recorded by MoCap while tracking the global path (pink) using only the PF method (14.4~m). (b) With dynamic obstacles: wheelchair trajectories in Experiment~A (Perpendicular Walk, left) and Experiment~B (Random Walk, right). The total distance traveled was 15.5~m (PF-RRT) and 17~m (PF-A*) in Experiment~A, and 15.7~m (PF-RRT) and 19.5~m (PF-A*) in Experiment~B. Pedestrian trajectories are omitted for clarity. Black rectangles are static rectangular obstacles and dashed orange rectangles are extended rectangles (buffer is 0.45~m).}
\label{fig:mocap}
\end{figure}

The global path used in the MoCap-based hardware experiments is shown in the left panel of Fig.~\ref{fig:mocap}a. The right panel of Fig.~\ref{fig:mocap}a shows the wheelchair trajectory obtained using only the PF method to follow this path when there are no dynamic obstacles in the environment. Fig.~\ref{fig:mocap}b presents the wheelchair trajectories for (A) the Perpendicular Walk experiment and (B) the Random Walk experiment. The total distance traveled by the wheelchair is larger for both TSP-PF-A* and TSP-PF-RRT than for the PF-only global path, which is due to the avoidance maneuvers performed when pedestrians obstruct the path. In Experiment A, the two methods travel approximately the same distance, with a difference of about 1.5~m. In Experiment B, TSP-PF-A* travels approximately 4~m more than TSP-PF-RRT. This is because, in the TSP-PF-A* run, a pedestrian blocked the wheelchair as it approached the first waypoint; the wheelchair therefore proceeded to the next point on the global path (a rectangle corner) and then maneuvered back to reach the first waypoint.

%Blue and red curves represent the trajectories obtained using TSP-PF-A* and TSP-PF-RRT, respectively.
%Consistent with the simulation results, RRT yielded a shorter total distance traveled than A* in both experiments.

% \begin{figure}[h]
% \centering
% \includegraphics[width=0.5\textwidth]{figures/mocap_exps.pdf}
% \caption{Wheelchair trajectories recorded by MoCap in the hardware experiments
% with dynamic obstacles. Left: Experiment~A (Perpendicular Walk), Right: Experiment~B (Random Walk). The total distance traveled was 15.5~m (RRT) and 17~m (A*) in Experiment~A, and 15.7~m (RRT) and 19.5~m (A*) in Experiment~B. Pedestrian trajectories are omitted for clarity.}\label{fig:path_mocap}
% \end{figure}

%5 adet PF-A* experiment, 
%5 adet PF-RRT experiment yapilacak.
% 1 adet sadece PF ile global path gezilecek.
% 10 insan.

% NOT: Iddiamiz A* ya da RRT iyi degil degil. Monte Carlo simulasyonlara gore ikisi de hemen hemen benzer bir performans gosteriyor gibi ve ikisyle de bu is yapilir. Bunlar PF uzerine ekleniyor zaten, onemli olan o. TSP ile once optimum path buluyor robot. PF o path'e cekiyor, iki tane force uretiyor bunun icin: tangential ve pulling force. Tek bir cekme degil, iki tane. Aksi halde, path'e dogru duzgun yaklasmaz ve mantikli path'ler ortaya cikmaz. Original katki bu.

\subsection{LiDAR- and RGB-D-Based Experiments}\label{sec7.2}

\begin{figure}[h]
\centering
\includegraphics[width=1.0\textwidth]{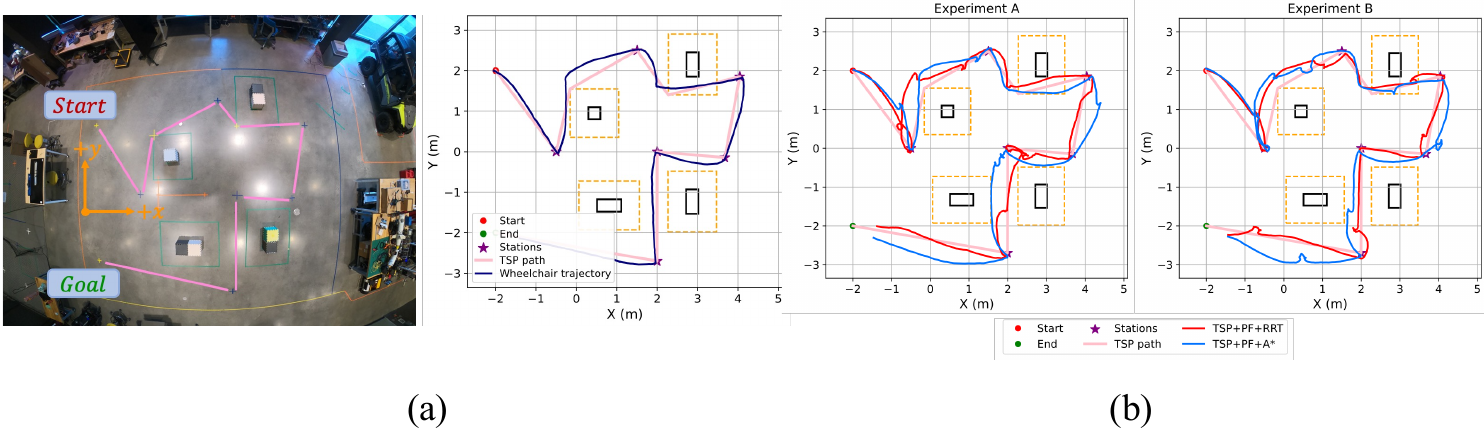}
\caption{LiDAR- and RGB-D-based hardware experiments. (a) Without dynamic obstacles. Left: Global path obtained by the TSP planner (19.6~m). Right: Wheelchair trajectory (navy) recorded by LiDAR while tracking the global path (pink) using only the PF method (19.9~m). (b) With dynamic obstacles: wheelchair trajectories in Experiment~A (Perpendicular Walk, left) and Experiment~B (Random Walk, right). The total distance traveled was 23.3~m (PF-RRT) and 22.3~m (PF-A*) in Experiment~A, and 22.4~m (PF-RRT) and 23.2~m (PF-A*) in Experiment~B. Pedestrian trajectories are omitted for clarity.}
\label{fig:lidar}
\end{figure}

In the outside-MoCap experiments, the 3D LiDAR is used for wheelchair localization, and the two RGB-D cameras at the front-left and front-right are used to detect and track the surrounding pedestrians (see Figs.~\ref{fig:physical_wheelchair} and~\ref{fig:general}). As in the MoCap-based experiments, the left panel of Fig.~\ref{fig:lidar}a shows the global path, and the right panel shows the trajectory of the wheelchair when only the PF method is used to follow this path, in the absence of dynamic obstacles. Fig.~\ref{fig:lidar}b shows the wheelchair trajectories for (A) the Perpendicular Walk and (B) the Random Walk experiments. As in Experiment A of the MoCap-based experiments, the total distance traveled by the wheelchair is approximately the same for TSP-PF-A* and TSP-PF-RRT in both Experiments A and B, with a difference of less than 1~m.

\subsubsection{Localization and SLAM Initialization}\label{sec7.2.1}

A single Livox Mid-360 LiDAR mounted at the rear of the wheelchair captures 3D point clouds of the environment at 10\,Hz. The point clouds are streamed to the onboard computer, where the FAST-LIO2 package~\cite{xu2022fastlio2} is used for odometry estimation by tightly coupling IMU measurements with the LiDAR point clouds through an Iterated Extended Kalman Filter (IEKF).

Since the magnetometer is not used due to electromagnetic interference in the test space, the LiDAR and Simultaneous Localization and Mapping (SLAM) stack are always started and initialized at the same position and orientation, so that all odometry is expressed relative to a common initial pose (see Fig.~\ref{fig:perception}). The LiDAR-based SLAM achieves a localization accuracy of approximately $\pm 4$\,cm, which is small compared to the 0.6\,m threshold used to determine whether the wheelchair has reached the next point along the global path and is therefore sufficient for the experiments. This accuracy was assessed by manually measuring the position of the wheelchair with a tape measure at known reference points in the test space and comparing it with the estimated pose.

% TODO: Consider adding a photo showing how the initialization pose is set up.

% A single Livox Mid-360 LiDAR mounted at the rear of the wheelchair captures 3D point clouds of the environment at 10\,Hz, producing approximately 200 000 points per second. The LiDAR is powered by a dedicated three-cell (3S) lithium-polymer (LiPo) battery supplying approximately 12\,V. The point clouds are streamed to the onboard computer over a wired Ethernet connection, where a SLAM pipeline estimates the pose of the wheelchair. We use the LIO-SAM package for odometry estimation, which tightly couples IMU measurements with the LiDAR point clouds.

% The test space is surrounded by ferrous materials and equipment that emits electromagnetic interference, so we do not rely on the magnetometer for heading estimation. Instead, the LiDAR and SLAM stack are always started and initialized at the same position and orientation within the test space, so that all odometry is expressed relative to a common initial pose. The test space is an open area of approximately $10 \times 10$\,m with abundant 3D structure, which provides the LiDAR with sufficient geometric features to estimate the pose with negligible drift. The perception stack is initialized once at the start of each session and then continuously publishes wheelchair odometry and pedestrian poses for several hours through a ROS~2 node. Under these conditions, the LiDAR-based SLAM achieves a localization accuracy of approximately $\pm 4$\,cm, which is sufficient for this experiment.

\subsubsection{Pedestrian Detection}\label{sec7.2.2}

\begin{figure}[h]
\centering
\includegraphics[width=0.5\textwidth]{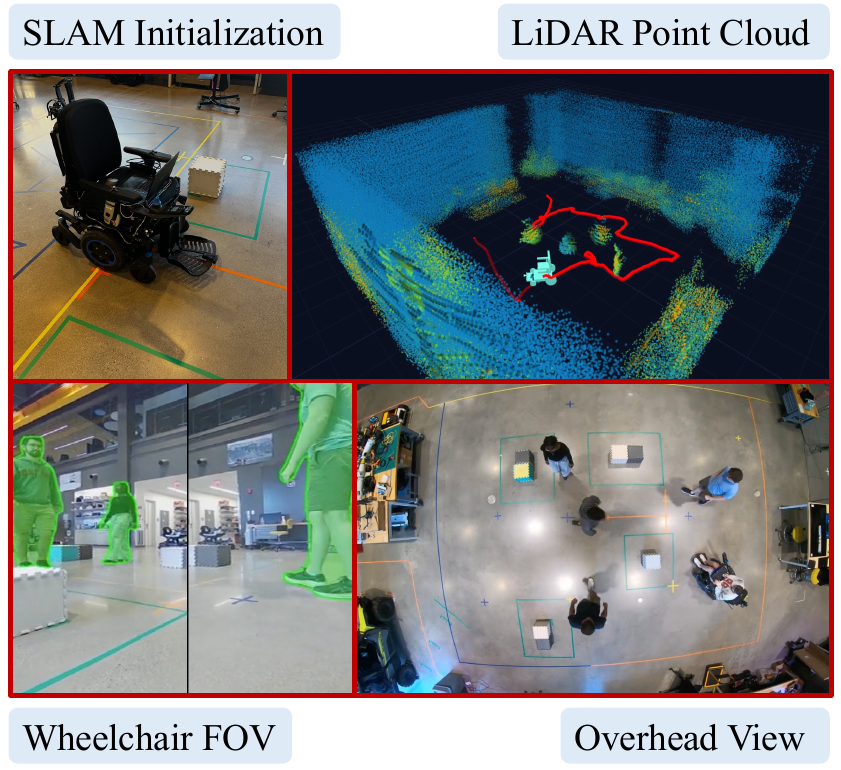}
\caption{Top left: The wheelchair is initialized at the same pose before each run, fixing the origin of the world frame. Top right: LiDAR point cloud with the estimated wheelchair trajectory (red). Bottom left: Pedestrian detection from the front-left and front-right RGB-D cameras (green: bounding boxes). Bottom right: Overhead view of the experiment.} 
\label{fig:perception}
\end{figure}

Two Orbbec Gemini RGB-D cameras are mounted at the front left and front right of the wheelchair, both facing forward. Together, they cover a field of view (FOV) of up to $45^\circ$ on each side of the wheelchair's heading (see Fig.~\ref{fig:perception}). Pedestrians are detected in the color frames of each camera independently, using a YOLO11x object detection model~\cite{yolo11_ultralytics} restricted to the person class. The 3D position of each detected pedestrian is computed from the depth image within its bounding box and transformed into the world frame using the wheelchair pose estimated by the LiDAR odometry. The pedestrian velocities are obtained from consecutive pedestrian positions.

The maximum pedestrian localization error in the world frame, measured with a tape measure at known reference points, is approximately $\pm 18$\,cm, while in most cases it remains within 10\,cm, which is sufficient for the experiments. This error includes the localization error of the wheelchair and is much smaller than the elliptical region used to model each pedestrian, whose semi-axes are at least $a_0 = 0.5$~m and $b_0 = 0.4$~m.

\section{Discussion}\label{sec8}
The simulation results show that the wheelchair model strongly influences the observed collision behavior. With the point model, TSP-PF-RRT avoided all collisions in the normal, crowded, and extra-crowded scenarios, whereas the collisions of TSP-PF-A* increased as the environment became more crowded. In all cases, both methods traveled comparable distances. With the more realistic elliptical model, which accounts for the physical dimensions of the wheelchair, TSP-PF-RRT also started to collide, but still less than TSP-PF-A*. In the normal scenario, both methods traveled almost the same distance, with 25\% fewer collisions for TSP-PF-RRT, and in the crowded scenario, TSP-PF-RRT traveled approximately 2\% more but had 33\% fewer collisions. These results suggest that, as the environment becomes more complex, TSP-PF-RRT achieves safer obstacle avoidance at the cost of a slightly longer path. The reported collision counts should be interpreted with the design of the simulation in mind. To test the limits of both planners and to determine which one outperforms the other, the simulated pedestrians deliberately approach and corner the wheelchair. Therefore, the collision counts primarily serve as a relative comparison of the two planners under extreme conditions. No collisions with dynamic obstacles were observed in the hardware experiments, and the TSP-PF-A* and TSP-PF-RRT planners performed nearly identically in terms of the total distance traveled. This may be partly because human pedestrians, as social agents, naturally and often unconsciously adjust their movements to avoid collisions. More importantly, the simulation environment is considerably more challenging than the hardware setting, both in the number of dynamic obstacles and in their adversarial behavior.

\textbf{Limitations:} First, the pedestrian motion in simulation is random and adversarial; while this is useful for testing the limits of the planners, it does not fully represent realistic human behavior in real environments. Second, hardware experiments were conducted in a laboratory environment rather than in a real assistive-living facility such as a nursing home. Third, only pedestrians were considered as dynamic obstacles; other wheelchairs, whose motion differs from that of pedestrians, were not included. Finally, the number of hardware trials was limited, and pedestrian localization relies on bounding-box detections, which introduce errors of up to 18~cm.

% SIMULATION'DA ROBOT VE ELLIPSE FARKI. A* VE RRT FARKI.

% No collisions with dynamic obstacles were observed in the hardware experiments. In both the van-ramp and the station-based experiments, the PF-A* and PF-RRT local planners performed nearly identically in terms of the total distance traveled by the wheelchair. This may be partly because human pedestrians, as social agents, naturally and often unconsciously adjust their movements to avoid collisions. More importantly, and this is the primary reason, the simulation environment is considerably more challenging than the hardware setting, both in the number of dynamic obstacles and in their behavior: the simulated pedestrians deliberately approach and corner the wheelchair. The simulation was designed this way to test the limits of both planners and to determine which one outperforms the other.

% LIMITATIONS: Insanlar mesela random ilerliyor simulasyonda, evet bu algoritmalarin limitlerini test etmek acisindan iyi oldu ama ne kadar gercekci bir insan davranisi oldugu tartisilir. Deneyin gercek bir nursing home setting'de yapilmamis olmasi ve lab'da yapilmasi da baska bir limitasyon. Sadece dynamic obstacle olarak insan var, wheelchair yoktu.

\section{Future Work and Conclusion}\label{sec9}

This study presented a hybrid multi-goal path planning framework for PWs that integrates TSP-based global task ordering, PF-guidance, and enhanced HLPs: RRT and A*. The framework was evaluated in simulations with point and elliptical wheelchair models under normal, crowded, and extra-crowded scenarios, and was implemented on a physical wheelchair using both MoCap and onboard LiDAR and RGB-D sensing. Both planners traveled comparable distances in all settings. In simulations, TSP-PF-RRT consistently had fewer collisions than TSP-PF-A*. No collisions were observed in the hardware experiments. These results indicate that the proposed framework enables collision-aware multi-goal navigation for PWs in dynamic environments, and that accounting for the wheelchair geometry is essential for a realistic evaluation of HLPs.

Future work will address the limitations identified above in four areas. First, real pedestrian motion data will be collected to develop more realistic human behavior models for simulation. Second, the optimal avoidance maneuver in two-agent wheelchair--pedestrian encounters will be investigated. Third, the framework will be extended to multiple PWs operating in the same environment. Finally, pedestrian localization will be improved by estimating each pedestrian's position from the 3D points of the segmentation mask rather than from the bounding box, which would reduce the effect of background pixels, body pose, and carried objects.

% BU MAKALEDE SU SU SU YAPILDI VE SUNLAR GOZLENDI. 

% Future work will also improve pedestrian localization by estimating each pedestrian's position from the 3D points of the segmentation mask rather than from the bounding box, which would reduce the effect of background pixels, body pose, and carried objects.

% Insanin daha gercekci modellenmesi icin data toplayacagiz. Ayni zamanda en optimal manevranin hesaplandigi bir makalemiz olacak.
% OPTIMAL CONTROL MAKALESI'NE GIRIS. 

% This work can be extended to a multiple agent PW. 

\backmatter

\bmhead{Supplementary Information}
The supplementary website, including the experiment videos, is available at \url{https://river-lab.github.io/AutonomousWheelchair/}.

\bmhead{Acknowledgements}

The authors would like to thank Adrian Patel, Aidan Reichenberg, Asher Hunter, Austin Allison and Wisena Joseph for their assistance during the hardware experiments, in particular for acting as pedestrians in the dynamic interaction trials. The authors also thank Northeastern University's Institute for Experiential Robotics (IER) for providing the experimental space and facilities.

\section*{Declarations}

\begin{itemize}
\item \textbf{Funding:} This research was funded, in part, by the Advanced Research Projects Agency for Health (ARPA-H) Agreement No. 140D042590012. The views and conclusions contained in this document are those of the authors and should not be interpreted as representing the official policies, either expressed or implied, of the U.S. Government. 
\item \textbf{Competing interests:} Ta\c{s}k{\i}n Pad{\i}r holds concurrent appointments as a Professor of Electrical and Computer Engineering at Northeastern University and as an Amazon Scholar. This work was conducted at Northeastern University's RIVeR Lab, led by Prof. Pad{\i}r, and is not associated with Amazon. The other authors declare no competing interests.
\item \textbf{Ethics approval and consent to participate:} Formal ethics approval was not required for this study, as the pedestrians in the hardware experiments were members of the RIVeR Lab who voluntarily participated, and no personal or sensitive data were collected. Informed consent was obtained from all individual participants included in the study.
\item \textbf{Consent for publication:} Informed consent for publication of identifiable images was obtained from all participants shown in the figures.
\item \textbf{Data availability:} The data supporting the findings of this study are available from the corresponding author upon reasonable request.
\item \textbf{Materials availability:} See Supplementary Information.
\item \textbf{Code availability:} \url{https://github.com/RIVeR-Lab/wheelchair-hybrid-path-planning}
\item \textbf{Author contributions:} D.L. developed the algorithms, performed the simulations and hardware experiments, prepared the supplementary website and video, and wrote the manuscript. E.O. contributed to the hardware experiments, the supplementary website and video, and to technical discussions. T.P. supervised the work and edited the manuscript. All authors read and approved the final manuscript.
\end{itemize}

\bibliography{sn-bibliography}% common bib file

@misc{leblebicioglu2026hybrid,
  title        = {A Hybrid Path Planning Framework for Assistive Wheelchair Navigation in Dynamic Pedestrian Environments},
  author       = {Leblebicio\u{g}lu, Damla and Barume, Rachel and Pad{\i}r, Ta\c{s}k{\i}n},
  year         = {2026},
  eprint       = {2610.42049},
  archivePrefix = {arXiv},
  primaryClass = {cs.RO},
  url          = {https://arxiv.org/abs/2610.42049}
}

@inproceedings{loke2026minimal,
  title={A Minimal, Deterministic Approach to Shared Control for Safe Powered Wheelchair Driving},
  author={Loke, Larisa Y\_C and Goh, Joel X and Puckett, Kyle S\_P and Zhang, Michelle H and Zaino, Nicole L and Argall, Brenna D},
  year={2026},
  organization={Proceedings of Robotics: Science and Systems (RSS)}
}

@article{cooper2008trends,
  title={Trends and issues in wheelchair technologies},
  author={Cooper, Rory A and Cooper, Rosemarie and Boninger, Michael L},
  journal={Assistive Technology},
  volume={20},
  number={2},
  pages={61--72},
  year={2008},
  publisher={Taylor \& Francis}
}

@article{simpson2008many,
  title={How many people would benefit from a smart wheelchair?},
  author={Simpson, Richard C and LoPresti, Edmund F and Cooper, Rory A},
  journal={Journal of Rehabilitation Research \& Development},
  volume={45},
  number={1},
  year={2008}
}

@article{jung2020path,
  title={Path planning algorithm for an autonomous electric wheelchair in hospitals},
  author={Jung, Yeontaek and Kim, Yoonjae and Lee, Woo Hyung and Bang, Moon Suk and Kim, Youdan and Kim, Sungwan},
  journal={IEEE Access},
  volume={8},
  pages={208199--208213},
  year={2020},
  publisher={IEEE}
}

@article{jian2021global,
  title={A global-local coupling two-stage path planning method for mobile robots},
  author={Jian, Zhiqiang and Zhang, Songyi and Chen, Shitao and Nan, Zhixiong and Zheng, Nanning},
  journal={IEEE Robotics and Automation Letters},
  volume={6},
  number={3},
  pages={5349--5356},
  year={2021},
  publisher={IEEE}
}

@article{yang2015improved,
  title={An improved ant colony optimization (I-ACO) method for the quasi travelling salesman problem (Quasi-TSP)},
  author={Yang, Jianyi and Ding, Ruifeng and Zhang, Yuan and Cong, Maoqin and Wang, Fei and Tang, Guoan},
  journal={International Journal of Geographical Information Science},
  volume={29},
  number={9},
  pages={1534--1551},
  year={2015},
  publisher={Taylor \& Francis}
}

@article{xie2018integrated,
  title={An integrated traveling salesman and coverage path planning problem for unmanned aircraft systems},
  author={Xie, Junfei and Carrillo, Luis Rodolfo Garcia and Jin, Lei},
  journal={IEEE control systems letters},
  volume={3},
  number={1},
  pages={67--72},
  year={2018},
  publisher={IEEE}
}

@article{khatib1986real,
  title={Real-time obstacle avoidance for manipulators and mobile robots},
  author={Khatib, Oussama},
  journal={The international journal of robotics research},
  volume={5},
  number={1},
  pages={90--98},
  year={1986},
  publisher={Sage Publications Sage CA: Thousand Oaks, CA}
}

@inproceedings{khatib1985real,
  title={Real-time obstacle avoidance for manipulators and mobile robots},
  author={Khatib, Oussama},
  booktitle={Proceedings. 1985 IEEE international conference on robotics and automation},
  volume={2},
  pages={500--505},
  year={1985}
}

@article{wang2020mobile,
  title={Mobile robot path planning in dynamic environments through globally guided reinforcement learning},
  author={Wang, Binyu and Liu, Zhe and Li, Qingbiao and Prorok, Amanda},
  journal={IEEE Robotics and Automation Letters},
  volume={5},
  number={4},
  pages={6932--6939},
  year={2020},
  publisher={IEEE}
}

@article{lu2025global,
  title={Global and local path planning of robots combining ACO and dynamic window algorithm},
  author={Lu, Yaping and Da, Chen},
  journal={Scientific Reports},
  volume={15},
  number={1},
  pages={9452},
  year={2025},
  publisher={Nature Publishing Group UK London}
}

@article{li2024path,
  title={Path planning and obstacle avoidance control of UUV based on an enhanced A* algorithm and MPC in dynamic environment},
  author={Li, Xiaohong and Yu, Shuanghe and Gao, Xiao-zhi and Yan, Yan and Zhao, Ying},
  journal={Ocean Engineering},
  volume={302},
  pages={117584},
  year={2024},
  publisher={Elsevier}
}

@article{lavalle2001randomized,
  title={Randomized kinodynamic planning},
  author={LaValle, Steven M and Kuffner Jr, James J},
  journal={The international journal of robotics research},
  volume={20},
  number={5},
  pages={378--400},
  year={2001},
  publisher={SAGE Publications}
}

@article{karaman2011sampling,
  title={Sampling-based algorithms for optimal motion planning},
  author={Karaman, Sertac and Frazzoli, Emilio},
  journal={The international journal of robotics research},
  volume={30},
  number={7},
  pages={846--894},
  year={2011},
  publisher={Sage Publications Sage UK: London, England}
}

@article{wang2020neural,
  title={Neural RRT*: Learning-based optimal path planning},
  author={Wang, Jiankun and Chi, Wenzheng and Li, Chenming and Wang, Chaoqun and Meng, Max Q-H},
  journal={IEEE transactions on automation science and engineering},
  volume={17},
  number={4},
  pages={1748--1758},
  year={2020},
  publisher={IEEE}
}

@article{guan2025improved,
  title={An improved artificial potential field with RRT star algorithm for autonomous vehicle path planning},
  author={Guan, Tian and Han, Yi and Kong, Michuang and Wang, Siyu and Feng, Deyuan and Yang, Wei},
  journal={Scientific reports},
  volume={15},
  number={1},
  pages={16982},
  year={2025},
  publisher={Nature Publishing Group UK London}
}

@article{rasekhipour2016potential,
  title={A potential field-based model predictive path-planning controller for autonomous road vehicles},
  author={Rasekhipour, Yadollah and Khajepour, Amir and Chen, Shih-Ken and Litkouhi, Bakhtiar},
  journal={IEEE Transactions on Intelligent Transportation Systems},
  volume={18},
  number={5},
  pages={1255--1267},
  year={2016},
  publisher={IEEE}
}

@article{huang2025self,
  title={A self-adaptive neighborhood search A-star algorithm for mobile robots global path planning},
  author={Huang, Jiabo and Chen, Chunmei and Shen, Junjie and Liu, Guihua and Xu, Feng},
  journal={Computers and Electrical Engineering},
  volume={123},
  pages={110018},
  year={2025},
  publisher={Elsevier}
}

@article{guo2020global,
  title={Global path planning and multi-objective path control for unmanned surface vehicle based on modified particle swarm optimization (PSO) algorithm},
  author={Guo, Xinghai and Ji, Mingjun and Zhao, Ziwei and Wen, Dusu and Zhang, Weidan},
  journal={Ocean Engineering},
  volume={216},
  pages={107693},
  year={2020},
  publisher={Elsevier}
}

@article{faigl2019fast,
  title={Fast heuristics for the 3-D multi-goal path planning based on the generalized traveling salesman problem with neighborhoods},
  author={Faigl, Jan and V{\'a}{\v{n}}a, Petr and Deckerov{\'a}, Jind{\v{r}}i{\v{s}}ka},
  journal={IEEE Robotics and Automation Letters},
  volume={4},
  number={3},
  pages={2439--2446},
  year={2019},
  publisher={IEEE}
}

@article{janovs2021multi,
  title={Multi-goal path planning using multiple random trees},
  author={Jano{\v{s}}, Jaroslav and Von{\'a}sek, Vojt{\v{e}}ch and P{\v{e}}ni{\v{c}}ka, Robert},
  journal={IEEE Robotics and Automation Letters},
  volume={6},
  number={2},
  pages={4201--4208},
  year={2021},
  publisher={IEEE}
}

@article{zhang2018path,
  title={Path planning of industrial robot based on improved RRT algorithm in complex environments},
  author={Zhang, Haojian and Wang, Yunkuan and Zheng, Jun and Yu, Junzhi},
  journal={IEEE Access},
  volume={6},
  pages={53296--53306},
  year={2018},
  publisher={IEEE}
}

@article{debnath2024integrated,
  title={An integrated geometric obstacle avoidance and genetic algorithm TSP model for UAV path planning},
  author={Debnath, Dipraj and Vanegas, Fernando and Boiteau, Sebastien and Gonzalez, Felipe},
  journal={Drones},
  volume={8},
  number={7},
  pages={302},
  year={2024},
  publisher={MDPI}
}

@misc{luci_sdk_docs,
  author       = {{LUCI Mobility, Inc.}},
  title        = {{LUCI} {ROS}~2 {SDK} Documentation},
  howpublished = {\url{https://lucimobility.github.io/luci-sdk-docs/}},
  year         = {2026},
  note         = {Accessed: Oct. 12, 2026}
}

@inproceedings{kuffner2000rrt,
  author    = {Kuffner, James J. and LaValle, Steven M.},
  title     = {{RRT}-Connect: An Efficient Approach to Single-Query Path Planning},
  booktitle = {Proceedings of the IEEE International Conference on Robotics and Automation (ICRA)},
  volume    = {2},
  pages     = {995--1001},
  year      = {2000}
}

@misc{rep103,
  author       = {Foote, Tully and Purvis, Mike},
  title        = {{REP} 103: Standard Units of Measure and Coordinate Conventions},
  year         = {2010},
  howpublished = {\url{https://www.ros.org/reps/rep-0103.html}},
  note         = {Accessed: Nov. 2026}
}

@article{xu2022fastlio2,
  author  = {Xu, Wei and Cai, Yixi and He, Dongjiao and Lin, Jiarong and Zhang, Fu},
  title   = {{FAST-LIO2}: Fast Direct {LiDAR}-Inertial Odometry},
  journal = {IEEE Transactions on Robotics},
  volume  = {38},
  number  = {4},
  pages   = {2053--2073},
  year    = {2022}
}

@misc{livox_mid360,
  author       = {{Livox Technology Company Limited}},
  title        = {Livox {Mid-360} {LiDAR}},
  howpublished = {\url{https://www.livoxtech.com/mid-360}},
  note         = {Accessed: Nov. 8, 2026}
}

@misc{orbbec_gemini336l,
  author       = {{Orbbec Inc.}},
  title        = {Orbbec {Gemini} {336L} Stereo Vision Camera},
  howpublished = {\url{https://www.orbbec.com/products/stereo-vision-camera/gemini-336l}},
  note         = {Accessed: Nov. 8, 2026}
}

@misc{yolo11_ultralytics,
  author       = {Jocher, Glenn and Qiu, Jing},
  title        = {Ultralytics {YOLO11}},
  year         = {2024},
  note         = {Version 11.0.0},
  howpublished = {\url{https://github.com/ultralytics/ultralytics}}
}

@article{sinyukov2014multi,
  title={Multi-modal control framework for a semi-autonomous wheelchair using modular sensor designs},
  author={Sinyukov, Dmitry and Desmond, Ross and Dickerman, Matthew and Fleming, James and Schaufeld, Jerome and Padir, Taskin},
  journal={Intelligent service robotics},
  volume={7},
  number={3},
  pages={145--155},
  year={2014},
  publisher={Springer}
}

@inproceedings{bhat2026lazy,
  title={Lazy Anytime Planning for the Dubins Moving Target Traveling Salesman Problem with Obstacles},
  author={Bhat, Anoop and Gutow, Geordan and Singh, Surya and Ren, Zhongqiang and Rathinam, Sivakumar and Choset, Howie},
  booktitle={2026 IEEE International Conference on Robotics and Automation (ICRA)},
  pages={17012--17019},
  year={2026},
}

@inproceedings{sinyukov2014augmenting,
  title={Augmenting a voice and facial expression control of a robotic wheelchair with assistive navigation},
  author={Sinyukov, Dmitry A and Li, Ran and Otero, Nicholas W and Gao, Runzi and Padir, Ta{\c{s}}k{\i}n},
  booktitle={2014 IEEE International Conference on Systems, Man, and Cybernetics (SMC)},
  pages={1088--1094},
  year={2014}
}

@article{bhat2026parallel,
  title={Parallel, asymptotically optimal algorithms for moving target traveling salesman problems},
  author={Bhat, Anoop and Gutow, Geordan and Vundurthy, Bhaskar and Ren, Zhongqiang and Rathinam, Sivakumar and Choset, Howie},
  journal={IEEE Transactions on Robotics},
  year={2026},
  volume  = {42},
  pages   = {2931--2951},
  publisher={IEEE}
}

@inproceedings{padir2015towards,
  title={Towards personalized smart wheelchairs: Lessons learned from discovery interviews},
  author={Pad{\i}r, Ta{\c{s}}k{\i}n},
  booktitle={2015 37th Annual International Conference of the IEEE Engineering in Medicine and Biology Society (EMBC)},
  pages={5016--5019},
  year={2015}
}

@article{greenberg2025effects,
  title={Effects of mobile robot passing-motion path curvature on human affective states in a hallway environment},
  author={Greenberg, Benjamin and Gonzalez-Bravo, Uriel and Yi, Jingang and Feldman, Jacob},
  journal={International Journal of Social Robotics},
  volume={17},
  number={12},
  pages={3103--3113},
  year={2025},
  publisher={Springer}
}

@article{schmuck2025sang,
  title={SANG: Socially aware navigation between groups},
  author={Schmuck, Viktor and Celiktutan, Oya},
  journal={International Journal of Social Robotics},
  volume={17},
  number={6},
  pages={989--1002},
  year={2025},
  publisher={Springer}
}

@article{camara2025proxemic,
  title={Modelling proxemic zone shapes from analytic and simulated motion strategies},
  author={Camara, Fanta and Fox, Charles},
  journal={International Journal of Social Robotics},
  volume={17},
  number={12},
  pages={3027--3045},
  year={2025},
  publisher={Springer}
}

@incollection{sinyukov2019wheelchairs,
  author    = {Sinyukov, Dmitry A. and Troy, Karen L. and Bowers, Matthew P. and Pad{\i}r, Ta{\c{s}}k{\i}n},
  title     = {Wheelchairs and Other Mobility Assistance},
  booktitle = {Biomechatronics},
  editor    = {Popovic, Marko B.},
  publisher = {Academic Press},
  address   = {London},
  year      = {2019},
  pages     = {373--417},
  doi       = {10.1016/B978-0-12-812939-5.00013-6}
}

@inproceedings{desmond2013development,
  title={Development of modular sensors for semi-autonomous wheelchairs},
  author={Desmond, Ross and Dickerman, Matthew and Fleming, James and Sinyukov, Dmitry and Schaufeld, Jerome and Padir, Taskin},
  booktitle={2013 IEEE conference on technologies for practical robot applications (TePRA)},
  pages={1--6},
  year={2013}
}
%% if required, the content of .bbl file can be included here once bbl is generated
%%\input sn-article.bbl

\end{document}